\documentclass{article}

\usepackage[final]{neurips_2022}

\usepackage[utf8]{inputenc} 
\usepackage[T1]{fontenc}    
\usepackage{hyperref}       
\usepackage{url}            
\usepackage{booktabs}       
\usepackage{amsfonts}       
\usepackage{nicefrac}       
\usepackage{microtype}      
\usepackage[dvipsnames]{xcolor}

\usepackage{graphicx}
\usepackage{subfigure}

\usepackage{amsmath}
\usepackage{amssymb}
\usepackage{mathtools}
\usepackage{amsthm}

\usepackage[textsize=tiny]{todonotes}
\usepackage[capitalize,noabbrev]{cleveref}
\usepackage{stmaryrd}
\usepackage{dsfont}
\usepackage{array}
\usepackage{caption}
\usepackage{enumitem}

\usepackage{wrapfig}

\theoremstyle{plain}
\newtheorem{theorem}{Theorem}[section]
\newtheorem{proposition}[theorem]{Proposition}
\newtheorem{lemma}[theorem]{Lemma}

\theoremstyle{definition}

\theoremstyle{remark}

\newcommand{\tcgr}[1]{\textcolor{gray}{#1}}

\newcommand{\argmin}{\operatornamewithlimits{argmin}}

\def\eps{\varepsilon}
\def\phi{\varphi}  

\def\T{\top}

\newcommand{\I}[1]{[#1]}
\newcommand{\sr}[1]{\operatorname{sr}(#1)}

\def\R{\mathbb{R}}

\newcommand{\F}{\mathcal{F}}

\newcommand{\N}{\mathbb{N}}

\newcommand{\NN}{\mathcal{N}}

\newcommand{\alg}{F}

\newcommand{\ssize}{n}
\newcommand{\idim}{m}
\newcommand{\odim}{d}
\newcommand{\rdim}{k}

\newcommand{\Sf}{S}

\newcommand{\G}{G}
\newcommand{\g}{g}

\newcommand{\Gr}{G_{\rdim}}

\newcommand{\Loss}{\mathcal{L}}
\newcommand{\loss}{l}

\newcommand{\Error}{\operatorname{Error}}
\newcommand{\Fr}{\text{Fr}}

\makeatletter
\newenvironment{tsubarray}[1]{%
  \vcenter\bgroup
  \Let@ \restore@math@cr \default@tag
  \baselineskip\fontdimen10 \scriptfont\tw@
  \advance\baselineskip\fontdimen12 \scriptfont\tw@
  \lineskip\thr@@\fontdimen8 \scriptfont\thr@@
  \lineskiplimit\lineskip
  \check@mathfonts
  \ialign\bgroup\ifx c#1\hfil\fi
    \normalfont\fontsize\sf@size\z@\selectfont\ignorespaces##\unskip\hfil\crcr
}{%
  \crcr\egroup\egroup
}
\makeatother
\newcommand{\tsub}[1]{\begin{tsubarray}{l}$#1$\end{tsubarray}}

\title{SketchBoost: Fast Gradient Boosted Decision Tree for Multioutput Problems}
\author{%
  Leonid Iosipoi\\
  Sber AI Lab and HSE University,
  Moscow, Russia \\
  \texttt{iosipoileonid@gmail.com} \\
  \And
  Anton Vakhrushev\\
  Sber AI Lab,
  Moscow, Russia \\
  \texttt{btbpanda@gmail.com} \\
}

\begin{document}

\maketitle

\begin{abstract}
Gradient Boosted Decision Tree (GBDT) is a widely-used machine learning algorithm that has been shown to achieve state-of-the-art results on many standard data science problems. We are interested in its application to multioutput problems when the output is highly multidimensional. Although there are highly effective GBDT implementations, their scalability to such problems is still unsatisfactory. In this paper, we propose novel methods aiming to accelerate the training process of GBDT in the multioutput scenario. The idea behind these methods lies in the approximate computation of a scoring function used to find the best split of decision trees. These methods are implemented in SketchBoost, which itself is integrated into our easily customizable Python-based GPU implementation of GBDT called Py-Boost. Our numerical study demonstrates that SketchBoost speeds up the training process of GBDT by up to over $40$ times while achieving comparable or even better performance.
\end{abstract}

\section{Introduction}
Gradient Boosted Decision Tree (GBDT) 
is one of the most powerful 
methods for solving prediction problems in both 
classification and regression domains. 
It is a dominant tool today 
in application domains
where tabular data is abundant, for example, in e-commerce, financial, and retail industries. 
GBDT has contributed to a large amount of top solutions in 
benchmark competitions such as Kaggle. 
This makes GBDT a fundamental component 
in the modern data scientist’s toolkit.
\par
\begin{wrapfigure}{r}{0.47\textwidth}
\centering
    \includegraphics[width=\linewidth]{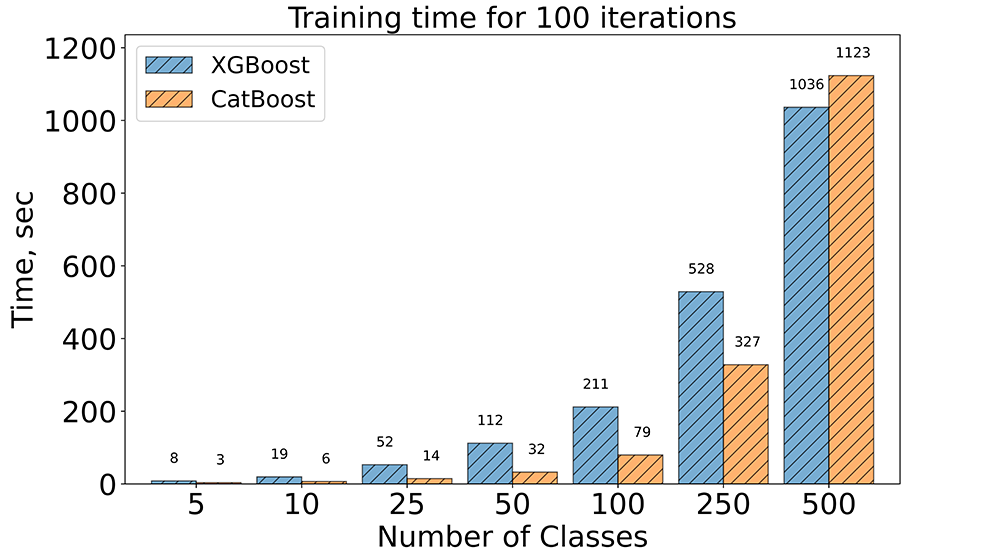}
    \caption{\small Training time of XGBoost and CatBoost for different number of classes on a synthetic dataset for multiclass classification. Synthetic dataset contains $2000$k instances, each described by $100$ features. The maximal tree depth was limited to $6$. The experiment was conducted on GPU.
    Further details are given in 
    \Cref{sec:synthetic_dataset}.
    \label{pic:train_time_main}}
    \vspace{-1em}
\end{wrapfigure}
The main focus of this paper is the scalability of GBDT to multioutput problems. Such problems include 
multiclass classification (a classification task with more than two mutually exclusive classes), multilabel classification (a classification task with more than two classes 
that are not mutually exclusive),  
and multioutput regression (a regression task with a multivariate response variable). These problems 
arise in various areas such as Finance
\citep*{oberman-waack-2016}, 
Multivariate Time Series Forecasting \citep*{zyz-2020},
Recommender Systems \citep*{jtl-2010}, and others.
\par
There are several extremely efficient, open-source, 
and production-ready implementations of gradient 
boosting such as 
XGBoost \citep{xgboost-2016}, 
LightGBM \citep*{lightgbm-2017},
and CatBoost~\citep*{catboost-2018}.
Even for them, learning a GBDT model for moderately 
large datasets can require much time.
Furthermore, this time also 
grows with the output size of a model.
\Cref{pic:train_time_main} demonstrates how 
rapidly the training time of XGBoost and
CatBoost grows with the output dimension.
Consequently, the number of possible applications of 
GBDT in the multioutput regime
is very limited.

\par
GBDT is a boosting-based algorithm 
that ensembles decision trees as \textquotedblleft base learners\textquotedblright. At each boosting step, 
a newly added tree improves 
the ensemble by minimizing 
the error of an already built composition.
There are two possible strategies on
how to use GBDT to handle a multioutput problem.
\par
\vspace{-0.5em}
\begin{itemize}[leftmargin=*]
    \item \textit{One-versus-all strategy.}
    Here, at each boosting step,
    a single decision tree is built for every output.
    Consequently, every output is handled separately.
    XGBoost and LightGBM use this strategy.
    \item \textit{Single-tree strategy.}
    Here, at each boosting step,
    a single multivariate decision tree is built 
    for all outputs.
    Consequently, all outputs are handled together.
    CatBoost uses this strategy.
\end{itemize}
\vspace{-0.5em}
\par
The computational complexity of both strategies is proportional to the number of outputs. Specifically, the one-versus-all strategy requires fitting a separate decision tree for each single output at each boosting step. The single-tree strategy requires scanning all the output dimensions (a)~to estimate the information gain 
during the search of the best tree structure
and (b)~to compute leaf values of 
a decision tree with a given structure 
(see details in \Cref{sec:preliminaries}).
A straightforward idea to reduce the training time of single-tree GBDT
is to exclude some of the outputs during the search of the tree structure which is the most time-consuming step of GBDT. However, this turns out to be rather challenging since it is unclear what outputs contribute the most to the information gain. In this paper, we address this problem and propose novel methods for fast scoring of multivariate decision trees which show a significant decrease in computational overhead without compromising the performance 
of the final model.
\par
\paragraph{Related work.}
Many suggestions have been made to speed up the
training process of GBDT. 
Some methods reduce
the number of data instances used 
to train each base learner.
For example, Stochastic Gradient Boosting (SGB) \citep{friedman-2002} 
chooses a random subset of data instances,
gradient-based one-side sampling (GOSS) 
\citep{lightgbm-2017}
keeps the instances with large gradients 
and randomly drops the instances with small gradients,
and
Minimal Variance Sampling (MVS)
\citep{mvs-2019} 
randomly chooses the instances 
to maximize the estimation accuracy of split scoring. 
Similarly, 
some methods reduce
the number of features. 
For example, one can 
choose a random subset of features 
or use principal component analysis or projection pursuit 
to exclude weak features; see
\citep{jl-1999, zhou-2012, afdp-2013}.
LightGBM~\citep{lightgbm-2017} uses exclusive feature 
bundling (EFB) where 
sparse features are greedily bundled together. 
CatBoost~\citep{catboost-2018} 
replaces categorical features with numerical ones 
using a special algorithm based on target statistics.
Finally, 
some methods reduce
the number of split candidates 
during the split scoring.
The pre-sorted algorithm~\citep{mrj-1996} 
enumerates all possible split 
points on the pre-sorted feature values. 
The histogram-based algorithm~\citep{ars-1998, rg-2003, lbw-2007} 
buckets continuous feature values into discrete 
bins and uses these bins to construct feature histograms.
\par
Regarding the multioutput regime, 
the existing methods to accelerate the training process
of GBDT naturally fall into the following two categories: 
problem transformation and algorithm adaptation. 
Transformation methods (see, for example, 
\citep{hklz2009,tl2012,kvj2012,cag2012,wtk2016}) 
reduce the number of targets before training a model.
They mainly differ in the choice of compression and decompression techniques 
and significantly rely on the problem structure or data assumptions. These methods pay a price in terms of prediction accuracy due to the loss of information during the compression phase, and as a result, they do not consistently outperform the full baseline. 
Adaptation methods directly extend some specific algorithms
to efficiently solve multioutput problems.
To the best of our knowledge, there are only 
two algorithm adaptation works for GBDT.
Namely,  
\citet{gbdt-sparse-2017} and \citet{gbdt-mo-2021}
consider models with sparse output and
discuss how to utilize this sparsity to enforce the 
leaf values to be also sparse.
Their modifications of GBDT are called
GBDT-Sparse and GBDT-MO (sparse). 
\par
We approach the problem 
of fast GBDT training in the multioutput regime
from a different perspective. 
Namely, instead of employing the model sparsity, we,
loosely speaking, approximate the scoring function 
used to find the best tree structure using
the most essential outputs 
while keeping other boosting steps without change.
The methods we suggest are 
completely different from the ones mentioned above
and can be applied to models 
with both dense and sparse outputs. 
Moreover, our methods can be easily combined with
transformation methods (by compressing the outputs beforehand 
and decompressing predictions afterward) or
the sparsity utilization as in GBDT-Sparse and GBDT-MO 
(by computing the optimal 
leaf values with sparsity constraint as in these algorithms). 
\paragraph{Contributions.}
The contributions of this work can be summarized as follows.
\par
\vspace{-0.5em}
\begin{itemize}[leftmargin=*]
    \item We propose and theoretically justify three novel methods 
    to speed up GBDT on multioutput tasks.
    These methods are generic, 
    they can be used with any loss function
    and do not rely on any specific data assumptions 
    (for example, sparsity or class hierarchy) 
    or the problem structure 
    (for example, multilabel or multiclass).
    Moreover, they 
    do not drop down the model quality and can be easily 
    integrated into any GBDT realization that uses 
    the single-tree strategy. 
    \item We implemented the proposed methods 
    in SketchBoost. SketchBoost itself is a part of
    our Python-based implementation of GBDT 
    called Py-Boost.
    This implementation seems to be of independent interest 
    since it does not use low-level programming languages 
    and is easily customizable. 
    Although it is written in Python, 
    it is fast since it works on GPU. 
    \item We present an empirical study using 
    public datasets which demonstrates that SketchBoost 
    achieves comparable or even better performance 
    compared to the existing state-of-the-art 
    boosting toolkits but in remarkably less time.
\end{itemize}
\vspace{-0.5em}
\par
\paragraph{Paper Organization.}
First, we review the GBDT algorithm in \Cref{sec:preliminaries}.
Next, we propose methods leading to a noticeable 
reduction in the training time of GBDT on multioutput tasks in 
\Cref{sec:methods}. 
We illustrate the performance of these methods on 
real-world datasets in \Cref{sec:experiments}.  
Proofs and experiment details are postponed 
to \Cref{sec:proofs} and \Cref{sec:experiment_details}.

\section{Preliminaries}
\label{sec:preliminaries}
Let $\{(x_i,y_i)\}_{i=1}^{\ssize}$ be a dataset with 
$\ssize$ samples,
where $x_i\in\R^{\idim}$ is an $\idim$ dimensional input and
$y_i\in\R^{\odim}$ is a $\odim$ dimensional output.
Let also $\F$ be a class of base learners, 
that is, functions $f:\R^{\idim}\to\R^{\odim}$.
In Gradient Boosting, the idea of which goes back 
to \citet{shapire-1990}, 
\citet{freund-1995}, \citet{freund-schapire-1997},
the model $\alg_T$ uses $T\in\N$ 
base learners $f\in\mathcal{F}$ 
and is trained in an additive and greedy manner.
Namely, at the $t$-th iteration, a newly 
added base learner $f$ improves the quality of 
an already built model $\alg_{t-1}$
by minimization of some specified loss function $\loss:\R^{\odim}\times\R^{\odim}\to\R$,
\[
  	 \Loss_{t}(f)  = \sum\nolimits_{i=1}^{\ssize} \loss(y_i, \alg_{t-1}(x_i) + f(x_i)).
\]
This optimization problem is usually approached by 
the Newton method using the second-order approximation
of the loss function
\begin{align}
    f^*_t  \in 
    \argmin_{f\in\F} \Biggl\{ 
        \sum_{i=1}^{\ssize} 
        \Bigl(
            g_i^{\T} f(x_i) 
		    + \frac12 \bigl(f(x_i)\bigr)^{\T} H_i f(x_i)
		\Bigr) 
   		+ \Omega(f) 
    \Biggr\},
    \label{eq:objfunc}
\end{align}
where we omitted a term independent of $f$; here
$\Omega(f)$ is a regularization term, 
usually added to build non-complex models, and
\begin{align}
		g_i = \nabla_a  \loss(y,a) 
		\Bigr|_{\tsub{y=y_i$\\$a=\alg_{t-1}(x_i)}}, 
		\quad
		H_i = \nabla^2_{aa} \loss(y,a) \Bigr|_{\tsub{y=y_i$\\$a=\alg_{t-1}(x_i)}}.
		\label{eq:derivatives}
\end{align}
Due to the complexity of optimization over
a large set of base learners $\F$, the problem 
\eqref{eq:objfunc} is solved typically in a greedy fashion  
which leads us to an approximate minimizer $f_t$.
Finally, the model $\alg_t$ is updated by applying 
a learning rate $\eps>0$ typically treated as a hyperparameter:
$
    \alg_{t} = \alg_{t-1} + \eps f_t.
$
\par
GBDT uses decision trees as the base learners $\F$;
see the seminal paper of \citet{friedman-2001}.
A decision tree is a model built by a recursive 
partition of the feature space into several 
disjoint regions.
Each final leaf is assigned to a value, 
which is a response of the tree in the given region.
Based on this construction mechanism,
a decision tree $f$ can be expressed as 
\[
    f(x) = \sum\nolimits_{j=1}^J v_j \cdot \I{x\in R_j},
\]  
where $\I{\text{predicate}}$ denotes the indicator function,
$J$ is the number of leaves, $R_j$ is the
$j$-th leaf, and $v_j\in\R^{\odim}$ is the value of $j$-th leaf.
The problem of learning $f_t$ can be naturally 
divided into two separate problems:
(1) finding the best tree structure 
(dividing the feature space into $J$ areas $R_1,\ldots,R_J$), 
and 
(2) fitting a decision tree with a given structure 
(computing leaf values $v_1,\ldots,v_J$).
\par
\paragraph{Finding the leaf values.} 
Since decision trees take constant values at each leaf,
for a decision tree $f_t$ with leaves $R_1,\ldots,R_{J}$,
we can optimize the objective function from \eqref{eq:objfunc}
for each leaf $R_j$ separately,
\begin{align*}
   		 v_j 
   		    &= \argmin_{v\in\R^{\odim}} \Biggl\{ \sum_{x_i\in R_j} \Bigl(g_i^{\T} v 
		    + \frac12 v^{\T} H_i v\Bigr) 
   		    + \frac{\lambda}{2}  \|v\|^2  \Biggr\} 
		    = - \Biggl(\sum_{x_i\in R_j} H_i + \lambda I\Biggr)^{-1} 
		        \Biggl(\sum_{x_i\in R_j}g_i \Biggr),
\end{align*}
where we employ $l_2$ regularization on leaf
values with a parameter $\lambda>0$; 
here $I$ denotes the identity matrix and 
$\|\cdot\|$ denotes the Euclidean norm. 
\par
It is worth mentioning that if the loss function
$\loss$ is separable with respect to different 
outputs, all Hessians $H_1,\ldots,H_{\ssize}$ 
are diagonal. If it is not the case,
it is a common practice to purposely simplify them 
to this extent in order to avoid time-consuming 
matrix inversion. 
It is done so in most of the single-tree GBDT algorithms
(for example, CatBoost, GBDT-Sparse, and GBDT-MO). 
We will also follow this idea in our work.
For diagonal Hessians, the optimal leaf values 
can be rewritten as
\begin{align}
	v_j 
	=
	-\frac{\sum_{i \in R} g_{i}^{j}}{\sum_{i \in R} h_{i}^{j}+\lambda},
	\quad
	\text{where }\,
	g_i = 
	\begin{pmatrix}
	g_i^1\\
	\vdots\\
	g_i^{\odim}
	\end{pmatrix}
	\,\text{ and }\,
	H_i = 
	\begin{pmatrix}
	h_i^1 & \ldots & 0\\
	\vdots & \ddots & \vdots \\
	0 & \ldots & h_i^{\odim}
	\end{pmatrix}.
    \label{eq:weightstaylorsimp}
\end{align}
\par
\paragraph{Finding the tree structure.} 
Substituting the leaf values 
from~\eqref{eq:weightstaylorsimp} 
back into the objective function, 
and omitting insignificant terms,
we obtain 
\begin{align}
    \operatorname{Loss}(f_t) = -\frac12 \sum_{j=1}^J \Sf(R_j),
    \quad\text{where}\quad
    \Sf(R)
	=
	\sum_{j=1}^d\frac{\bigl( \sum_{x_i \in R} g_{i}^{j} \bigr)^2}{\sum_{x_i \in R} h_{i}^{j}+\lambda}.
    \label{eq:scorefunc}
\end{align}
The function $\Sf(\cdot)$ will be referred to as the scoring 
function. 
To find the best tree structure,
we use a greedy algorithm that starts from a
single leaf and iteratively adds branches 
to the tree. At a general step, 
we want to split one of existing leaves.
To do this, we iterate through all leaves, 
features, and thresholds for each feature
(they are usually determined by the histogram-based algorithm).
For all leaves $R$ and all possible 
splits for $R$, say $R_{\text{left}}$ and $R_{\text{right}}$,
we compute the impurity score given by
$
	\Sf(R_{\text{left}}) + \Sf(R_{\text{right}}).
$
The best split is considered the one 
which achieves the largest impurity score. 
This is equivalent to maximization of the information gain 
which is usually defined as the difference between values of the
loss function before 
and after the split, that is,
\[
    \text{Gain} = -0.5\Bigl(\Sf(R) - \bigl(\Sf(R_{\text{left}}) + \Sf(R_{\text{right}})\bigr) \Bigr).
\]
\par
Similar to the previous step, 
some simplifications can be made to 
speed up computation of the scoring 
function which is done a tremendous number of times. 
For instance, GBDT-Sparse 
does not use the second-order information at all 
(Hessians are simplified to identity matrices). 
In the multioutput regime of CatBoost, 
the second-order derivatives are left out during 
the split search and are used only to compute leaf values.
GBDT-MO uses the second-order
derivatives in both steps
but it increases the computational complexity twice 
(histograms for both gradients and Hessians
need to be accumulated).

\section{Sketched Split Scoring}
\label{sec:methods}
In this section, we propose three novel methods
to speed up the split search
for multivariate decision trees. 
These methods can achieve a good balance
between reducing the computational complexity in
the output dimension and keeping the accuracy for 
learned decision trees.
They are generic and can 
be used together with the methods mentioned 
in the Related work section that
aim at reducing the number of sample instances, 
features, or split candidates.
Moreover, the proposed methods 
are easy to implement 
upon modern boosting frameworks 
such as XGBoost, LightGBM, and CatBoost.
\par
As it was mentioned before, there are two 
\textquotedblleft best practices\textquotedblright\ 
to speed up the training of a GBDT model 
on multioutput tasks:
(a)~to totally ignore the
second-order derivatives during the split search
and (b)~to use only the main diagonal of the
second-order derivatives 
to compute the leaf values. 
It is done so, for example, in CatBoost, 
one of the few boosting toolkits that
use the single-tree strategy and 
achieve state-of-the-art results on 
multioutput problems.
We will also develop our work on this basis.
\par
The proposed methods are applied at each boosting step before the search for the best tree structure and after first- and second-order derivatives
(see \eqref{eq:derivatives}) are computed. 
The key idea of the proposed methods 
is to reduce the number of gradient values used
in the split search 
so that the scoring function $\Sf$ from~\eqref{eq:scorefunc} or, equivalently,
the information gain will not change much.
Specifically, the scoring function 
without the second-order 
information can be rewritten as 
\[
    \Sf_{\G}(R) = \frac{\bigl\|\G^{\T}v_{R}\bigr\|^2}{|R|+ \lambda},
    \quad\text{where}\ \,
    \G =
    \begin{pmatrix}
    g_1^1 & g_1^2 & \ldots & g_1^{\odim}\\
    \vdots &  \vdots & \ddots & \vdots\\
    g_{\ssize}^1 & g_{\ssize}^2 & \ldots & g_{\ssize}^{\odim}\\
    \end{pmatrix}
    \ \,\text{and}\ \,
    v_{R} = 
    \begin{pmatrix}
    \I{x_1 \in R} \\
    \vdots \\
    \I{x_{\ssize} \in R} \\
    \end{pmatrix}.
\]
Here $\G\in\R^{\ssize \times \odim}$ is the gradient matrix 
and $v_{R}$ is the indicator vector of the leaf $R$ 
(its $i$-th coordinate is equal to $1$ if $x_i\in R$ and $0$ otherwise). 
Note that we added the subscript to $\Sf$ to indicate its 
dependence on the gradient matrix~$G$.
To reduce the complexity of computing $\Sf_{\G}$ in $\odim$,
we approximate it with $\Sf_{\Gr}$ 
for some other matrix $\Gr\in\R^{\ssize \times \rdim}$ 
with $\rdim \ll \odim$. 
We will refer to $\Gr$ as the sketch matrix
and to $\rdim$ as the reduced dimension or sketching dimension.
We emphasize that $\Gr$ 
is assumed to be used only in 
building histograms and finding the tree structure. 
After this, the optimal leaf values 
of a tree 
are assumed to be computed fairly using the
full gradient matrix $\G$.
\par
Further we discuss three novel methods
to construct reasonably good sketches $\Gr$ ---
Top Outputs, Random Sampling, and Random Projections.
These methods are motivated by the minimization
of the approximation error given by
\begin{align*}
    \Error(\Sf_{\G},\Sf_{\Gr}) 
    = \sup\nolimits_{R}\bigl|\Sf_{\G}(R) - \Sf_{\Gr}(R)\bigr|.
    \label{eq:errorfunc}
\end{align*}
Here the supremum is taken over all
possible leaves $R$. 
The reason for this choice is that 
we want the proposed approximation to be 
universal and uniformly accurate for all 
splits we will possibly iterate over. 
In \Cref{sec:proofs},
we show that the proposed methods lead
to a nearly-optimal upper bounds on the proposed error.
Since the corresponding optimization problem 
is an instance of Integer Programming problem,
methods leading to the optimal 
upper bounds 
can be obtained only by brute force,
which is not an option in our case.
For further details see \Cref{sec:proofs}.

\subsection{Top Outputs}
\label{sec:topoutputs}
The key idea of Top Outputs is rather straightforward:
to choose the columns of $\G$ with the largest Euclidian norm.
Namely, by a slight abuse of notation, let us denote the columns of $\G$ by $\g_1,\ldots,\g_{\odim}$.
Let also $i_1,\ldots,i_d$ be the 
indexes which sort the
columns of $\G$ in descending order by their norm, 
that is, 
$
    \|\g_{i_1}\| 
    \ge \|\g_{i_2}\| 
    \ge \ldots 
    \ge \|\g_{i_\odim}\|
$. 
Now the full gradient matrix and its sketch
can be written as
\begin{align*}
    \G =
    \begin{pmatrix}
    \vert & \vert & & \vert\\
    \g_1 & \g_2 & \ldots & \g_{\odim}\\
    \vert & \vert & & \vert\\
    \end{pmatrix}
    \quad\text{and}\quad
    \Gr =
    \begin{pmatrix}
    \vert & \vert & & \vert\\
    \g_{i_1}& \g_{i_2} & \ldots & \g_{i_\rdim}\\
    \vert & \vert & & \vert\\
    \end{pmatrix}.
\end{align*}
The parameter $\rdim$ here can be chosen adaptively 
to the norms of $\g_1,\ldots,\g_{\odim}$. 
We have not considered 
this generalization here since, in our view, 
it will greatly complicate the algorithm. 
Moreover, the adaptive choice of $\rdim$ 
may result in large values for this parameter 
and hence less gain in training time.
\par
It is worth pointing out that Top Outputs 
is akin to the Gradient-based One-Side Sampling (GOSS),
which is successively used in LightGBM; see \citep{lightgbm-2017}.
In GOSS, data instances with small gradients are excluded
to speed up the split search. Similarly,
Top Outputs excludes output components with small gradient values. 
\par
This method has one major drawback. 
This method chooses top $\rdim$ output dimensions
which may not vary much from step to step.
For instance, if several columns have large norms 
and others have medium norms, Top Outputs 
may completely ignore the latter columns during 
the split search.
Below
we consider another method that deals with this 
problem by introducing the randomness in the choice
of output dimensions.

\subsection{Random Sampling}
\label{sec:randsamp}
The probabilistic approach 
for algebraic computations, sometimes called the
\textquotedblleft Monte-Carlo method\textquotedblright, 
is ubiquitous; we refer the reader to the monographs of
\citet{robert-casella-2005},\citet{mahoney-11}, and \citet{wodruff-2014}.
Here we consider its application
to the fast split search.
\par
The key idea of Random Sampling is to randomly sample 
the columns of $\G$ with probabilities proportional
to their norms.
Namely, we define the sampling probabilities by
\begin{align*}
    p_i = \|\g_i\|^2 \Big/\sum\nolimits_{j=1}^{\odim}\|\g_j\|^2,
    \quad
    i=1,\ldots,\odim.
\end{align*}
These probabilities are known to be optimal for random sampling since they minimize the variance of the resulting estimate; see, for example, \citep{robert-casella-2005}.
Further, let ${i}_1,\ldots,{i}_{\rdim}$ be 
independent and identically distributed random variables 
taking values $j$ with probabilities $p_j$, 
$j=1,\ldots,\rdim$. These random variables represent indexes 
of the chosen columns of $\G$. 
Finally, we consider the following sketch
\[
    \Gr =
    \begin{pmatrix}
    \vert & \vert & & \vert\\
    \overline{\g}_{i_1} & \overline{\g}_{i_2} & \ldots & \overline{\g}_{i_\rdim}\\
    \vert & \vert & & \vert
    \end{pmatrix},
    \quad\text{where}\quad
    \overline{\g}_{i} = \frac{1}{\sqrt{\rdim p_i}} \, \g_{i}.
\]
The additional column normalization 
by $1/\sqrt{\rdim p_i}$ is needed
for unbiasedness of the resulting estimate. 
\par
\par
There is a close affinity between 
Importance Sampling and 
Minimal Variance Sampling (MVS) of \citet{mvs-2019}.
MVS decreases the number of sample instances in the split 
search by maximizing 
the estimation accuracy of split scoring.
Our idea is the same with the only difference that 
it is applied to output dimensions rather than sample instances.
\par
Random Sampling works well especially in the extreme cases
as those mentioned above. 
For example, if several outputs have large 
weights and others have medium weights, 
Random Sampling will not ignore the latter 
outputs due to randomness. Or, if the number 
of outputs with large weights is larger than $\rdim$, 
Random Sampling will choose different 
output dimensions at different steps. 
As a result, the corresponding base learners will 
also be quite different, which usually leads 
to a better generalization ability of the ensemble; see \cite{breiman1996bagging}.

\subsection{Random Projections}
\label{sec:randproj}
In the previous section, the sketch $\Gr$
was constructed by sampling columns
from $\G$ according to some probability distribution. 
This process can be viewed as 
multiplication of $\G$ by a matrix $\Pi$,
$
    \Gr = \G\Pi,
$
where $\Pi\in\R^{\odim\times\rdim}$ 
has independent columns, 
and each column is all zero except for a $1$ 
in a random location. 
In Random Projections, 
we consider sampling matrices $\Pi$,
every entry of which is an independently sampled 
random variable.
This results in using random 
linear combinations of columns of 
$\G$ as columns of $\Gr$.
\par
This approach is based on the 
Johnson-Lindenstrauss (JL) lemma; see 
the seminal paper of \citet{johnson-lindenstrauss-1984}.
They showed that projections $\Pi$ from $\odim$ dimensions onto a
randomly chosen $\rdim$-dimensional subspace do not 
distort the pairwise distances too much.
\citet{indyk-motwani-1998} 
proved that to obtain the same guarantee, 
one can independently sample every entry of $\Pi$ 
using the normal distribution.
In fact, this is true for many other distributions; 
see, for example, \citet{achlioptas-2003}.
Since there was no significant difference 
between distributions in our numerical experiments,
we decided to focus on the normal distribution.
\par
In Random Projections, we consider the following sketch
\[
    \Gr = \G\Pi,
\]
where $\Pi\in\R^{\odim\times\rdim}$ is a random matrix filled with independently sampled $\NN(0,\rdim^{-1})$
entries.
In \Cref{sec:proofs}, we discuss why this choice leads to a nearly-optimal solution to the problem we consider and why the property of preserving the pairwise distances matters here. 
\par
Random Projections has the same merits as Random Sampling
since it is also a random approach.
Besides that, the sketch matrix $\Gr$ here uses
gradient information from all outputs
since each column of $\Gr$ is a linear combination 
of columns of $\G$.

\subsection{Complexity analysis.} 
Most of the GBDT frameworks use histogram-based algorithm to speed up split finding; see \citep*{ars-1998}, \citep*{jin-agrawal-2003}, and \citep*{lwb-2008}.
Instead of finding the split points on  
all possible 
feature values, 
histogram-based algorithm buckets feature values 
into discrete bins and uses these bins 
to construct feature histograms during training. 
Let us say that the number of possible splits 
per feature is limited to $h \ll \ssize$ 
(usually $h\leq256$ to store the histogram bin index using a single byte).
It is shown in \citep{lightgbm-2017} that 
in the case of a single output, splitting a leaf $R$ 
with $n_R$ samples requires $O(\idim n_R)$ operations 
for histogram building and $O(h\idim)$ operations for split finding.
As a result, if the actual tree construction 
is performed using a depth-first-search 
algorithm, 
the complexity of building a complete 
tree of depth $D$ is $O(D \ssize\idim +2^Dh\idim)$.
In the multioutput scenario,
this complexity increases by $\odim$ times: 
splitting a leaf $R$ with $n_R$ samples costs 
$O(\idim n_R \odim+h\idim\odim)$ and
depth-wise tree construction costs 
$O(D\idim \ssize \odim+2^Dh\idim\odim)$. 
The methods we propose reduce the 
impact of $\odim$ to $\rdim$ with $\rdim \ll \odim$.
They require a preprocessing step which 
can be done, depending on the method,
in $O(\ssize \odim \rdim)$ or $O(\ssize \odim)$ operations.
As a result, the complexity of building 
a complete tree of depth $D$
using the depth-first search 
can be reduced from
$O(D\idim \ssize \odim +2^Dh\idim\odim)$ to $O(\ssize \odim  + D\idim \ssize \rdim +2^Dh\idim \rdim)$.
Taking into account that $\ssize$, $\idim$, and $\odim$ 
can be extremely large,
these methods may lead to a significant improvement 
in the training time.


\section{Numerical Experiments}
\label{sec:experiments}
In this section, we numerically compare 
(a) the proposed methods from \Cref{sec:methods} to speed up
GBDT in the multioutput regime 
and (b) 
existing state-of-art boosting toolkits supporting
multioutput tasks. 
\par 
\vspace{-0.5em}
\paragraph{Data.}
The experiments are conducted on $9$ real-world 
publicly available datasets from Kaggle, OpenML, and
\href{http://mulan.sourceforge.net/datasets.html}{Mulan}\footnote{\url{http://mulan.sourceforge.net/datasets.html}}
for multiclass ($4$ datasets) and multilabel ($3$ datasets) classification and multitask regression ($2$ datasets).
The associated details are given in 
\Cref{tb:datasets} in \Cref{sec:experiment_details}.
\par
\vspace{-0.5em}
\paragraph{Py-Boost.}
We implemented a simple and fast GBDT 
toolkit called {Py-Boost}. 
It is written in Python and hence is easily 
customizable.
Py-Boost works only on GPU 
and uses Python GPU libraries such as CuPy and Numba.
It follows the classic scheme described in 
\citep{xgboost-2016}; further details are provided 
in \Cref{sec:about_pyboost}.
Py-Boost is available on 
\href{https://github.com/sb-ai-lab/Py-Boost}{GitHub}\footnote{\url{https://github.com/sb-ai-lab/Py-Boost}}.
\par
\vspace{-0.5em}
\paragraph{SketchBoost.} 
SketchBoost is a part of Py-Boost library which implements the following three sketching strategies for fast split search: 
\textbf{Top Outputs} (\Cref{sec:topoutputs}),
\textbf{Random Sampling} (\Cref{sec:randsamp}),
and \textbf{Random Projections} (\Cref{sec:randproj}).
For convenience, Py-Boost without any sketching strategy
is referred to as \textbf{SketchBoost Full}.
All the following experimental results 
and evaluation code are also available on 
\href{https://github.com/sb-ai-lab/sketchboost-paper}{GitHub}\footnote{\url{https://github.com/sb-ai-lab/SketchBoost-paper}}. 
\par
\vspace{-0.5em}
\paragraph{Baselines.} Primarily we compare SketchBoost with 
    \textbf{XGBoost} (v1.6.0)
and
    \textbf{CatBoost} (v1.0.5).
There are two reasons why we have chosen 
these GBDT frameworks. 
First, 
they are commonly used among
practitioners and represent two different approaches 
to multiouput tasks (one-vs-all and single-tree). 
Second, they can be efficiently trained 
on GPU, 
which allows us to compare their training time 
with GPU-based SketchBoost (with an exception for CatBoost which 
supports 
multilabel classification and multioutput regression 
tasks only on CPU).
The reason why we have not considered LightGBM as a baseline is that 
it uses the same multiouput strategy as XGBoost (one-vs-all) and
its latest version (v3.3.2) does not
support multilabel classification 
and multioutput regression tasks
without external wrappers.
Further, we also compare 
SketchBoost with
\textbf{TabNet} (v3.1.1), a popular deep learning model
for tabular data; see \citep*{arik2021tabnet}.
Our aim here is not to make an exhaustive comparison with 
existing deep learning approaches (it deserves its own investigation), but to make a comparison with a 
different in nature
approach which moreover often has satisfactory complexity on 
large multioutput datasets.
\par
\vspace{-0.5em}
\paragraph{Experiment Design.} 
If there is no official train/test split,
we randomly split the data into training and 
test sets with ratio $80\%$-$20\%$. 
Then each algorithm is trained 
with 5-fold cross-validation 
(the train folds are used to fit a model and 
the validation fold is used for early stopping).
We evaluate all the obtained models
on the test set
and get $5$ scores for each model. 
The overall performance of algorithms is 
computed as an average score.
As a performance measure,
we use the cross-entropy for classification 
and RMSE for regression, but,
for the sake of completeness, 
we also report the accuracy score for classification and R-squared score for regression
in \Cref{sec:experiment_results}. 
For XGBoost, Catboost, and TabNet, we do 
the hyperparameter optimization using the Optuna 
framework \citep*{akiba2019optuna}. 
For SketchBoost, we use the same 
hyperparameters as for CatBoost 
(to speed up the experiment; we do not expect that 
hyperparameters will vary much since we
use the same single-tree approach).
The sketch size $\rdim$ is iterated
through the grid $\{1, 2, 5, 10, 20\}$
(or through a subset of this grid with values less than the output dimension).
Further information on experiment design is given in 
\Cref{sec:experiment_design,sec:experiment_hype,sec:experiment_design_tabnet}.
\par

\begin{table}[ht!]
  \setlength\tabcolsep{2pt}
  \setlength\extrarowheight{2pt}
  \captionsetup{justification=centering}
  \centering
  \vspace{-1em}
  \caption{\small
  Test errors (cross-entropy for classification and RMSE for regression)
  $\pm$ their standard deviation.
  \label{tb:test_score}}
  \vspace{0.2em}
  \scalebox{0.63}{
  \begin{tabular}{@{\extracolsep{4pt}}lccccccc@{}}
    \toprule
    & \multicolumn{4}{c}{\textbf{SketchBoost}} & \multicolumn{3}{c}{\textbf{Baseline}} 
    \vspace{0.4em}\\
    \cline{2-5}  \cline{6-8} \vspace{-0.2em}
    \textbf{Dataset} & \textbf{Top Outputs} & \textbf{Random Sampling} & \textbf{Random Projection} & \textbf{SketchBoost Full} & \textbf{CatBoost} & \textbf{XGBoost} & \textbf{TabNet} \\
     &\small{(for the best $k$)} & \small{(for the best $k$)} & \small{(for the best $k$)} & \small{(multioutput)} & \small{(multioutput)} & \small{(one-vs-all)} & \small{(multioutput)}\\
    \midrule
    \textbf{Multiclass classification} & & & & & \textbf{} &\textbf{} & \textbf{} \\
    Otto (9 classes)& 0.4715 & 0.4636 & \textbf{0.4566} & 0.4697 & 0.4658 & 0.4599 & 0.5363\\
    & \qquad\tcgr{$\pm$0.0035} & \qquad\tcgr{$\pm$0.0026} & \qquad\tcgr{$\pm$0.0023} & \qquad\tcgr{$\pm$0.0030} & \qquad\tcgr{$\pm$0.0033} & \qquad\tcgr{$\pm$0.0028} & \qquad\tcgr{$\pm$0.0063}\\
    SF-Crime (39 classes) & 2.2070 & 2.2037 & 2.2038 & 2.2067 & \textbf{2.2036} & 2.2208 & 2.4819\\
    & \qquad\tcgr{$\pm$0.0005} & \qquad\tcgr{$\pm$0.0004} & \qquad\tcgr{$\pm$0.0004} & \qquad\tcgr{$\pm$0.0003} & \qquad\tcgr{$\pm$0.0005} & \qquad\tcgr{$\pm$0.0008} & \qquad\tcgr{$\pm$0.0199}\\
    Helena (100 classes) & 2.5923 & 2.5693 & \textbf{2.5673} & 2.5865 & 2.5698 & 2.5889\ & 2.7197\\
    & \qquad\tcgr{$\pm$0.0024} & \qquad\tcgr{$\pm$0.0022} & \qquad\tcgr{$\pm$0.0026} & \qquad\tcgr{$\pm$0.0025} & \qquad\tcgr{$\pm$0.0025} & \qquad\tcgr{$\pm$0.0032} & \qquad\tcgr{$\pm$0.0235}\\
    Dionis (355 classes) & 0.3146 & 0.3040 & \textbf{0.2848} & 0.3114 & 0.3085 & 0.3502 & 0.4753\\
    & \qquad\tcgr{$\pm$0.0011} & \qquad\tcgr{$\pm$0.0014} & \qquad\tcgr{$\pm$0.0012} & \qquad\tcgr{$\pm$0.0009} & \qquad\tcgr{$\pm$0.0010} & \qquad\tcgr{$\pm$0.0020} & \qquad\tcgr{$\pm$0.0126}\\
    \midrule
    \textbf{Multilabel classification} &  & & & & \textbf{} & & \\
    Mediamill (101 labels) & 0.0745 & 0.0745 & \textbf{0.0743} & 0.0747 & 0.0754 & 0.0758 & 0.0859\\
    & \qquad\tcgr{$\pm$1.3e-04} & \qquad\tcgr{$\pm$1.3e-04} & \qquad\tcgr{$\pm$1.1e-04} & \qquad\tcgr{$\pm$1.3e-04} & \qquad\tcgr{$\pm$1.1e-04} & \qquad\tcgr{$\pm$1.1e-04} & \qquad\tcgr{$\pm$3.3e-03}\\
    MoA (206 labels) & 0.0163 & 0.0160 & \textbf{0.0160} & 0.0160 & 0.0161 & 0.0166 & 0.0193\\
    & \qquad\tcgr{$\pm$2.2e-05} & \qquad\tcgr{$\pm$1.0e-05} & \qquad\tcgr{$\pm$6.0e-06} & \qquad\tcgr{$\pm$9.0e-06} & \qquad\tcgr{$\pm$2.6e-05} & \qquad\tcgr{$\pm$2.1e-05} & \qquad\tcgr{$\pm$3.0e-04}	\\
    Delicious (983 labels) & 0.0622 & 0.0619 & 0.0620 & 0.0619 & \textbf{0.0614} & 0.0620 & 0.0664\\
    & \qquad\tcgr{$\pm$6.2e-05} & \qquad\tcgr{$\pm$5.9e-05} & \qquad\tcgr{$\pm$6.2e-05} & \qquad\tcgr{$\pm$5.5e-05} & \qquad\tcgr{$\pm$5.2e-05} & \qquad\tcgr{$\pm$3.3e-05} & \qquad\tcgr{$\pm$8.0e-04}\\
    \midrule
    \textbf{Multitask regression} &  & & & & \textbf{} & \\
    RF1 (8 tasks) & 1.1860 & 0.9944 & 0.9056 & 1.1687 & \textbf{0.8975} & 0.9250 & 3.7948\\
    & \qquad\tcgr{$\pm$0.1366} & \qquad\tcgr{$\pm$0.1015} & \qquad\tcgr{$\pm$0.0582} & \qquad\tcgr{$\pm$0.0835} & \qquad\tcgr{$\pm$0.0384} & \qquad\tcgr{$\pm$0.0307} & \qquad\tcgr{$\pm$1.5935}\\
    SCM20D (16 tasks) & 88.7442 & 86.2964 & \textbf{85.8061} & 91.0142 & 90.9814 & 89.1045 & 87.3655\\
    & \qquad\tcgr{$\pm$0.6346} & \qquad\tcgr{$\pm$0.4398} & \qquad\tcgr{$\pm$0.5534} & \qquad\tcgr{$\pm$0.3397} & \qquad\tcgr{$\pm$0.3652} & \qquad\tcgr{$\pm$0.4950} & \qquad\tcgr{$\pm$1.3316}\\
    \bottomrule
  \end{tabular}}
  \vspace{-0.5em}
\end{table}

\begin{table}[ht!]
  \setlength\tabcolsep{3pt}
  \setlength\extrarowheight{2pt}
  \captionsetup{justification=centering}
  \centering
  \vspace{-1em}
  \caption{\small
  Training time per fold in seconds.
  \\(CatBoost does not support multilabel classification 
    and multioutput regression tasks in the GPU mode.)\label{tb:train_time}}
  \vspace{0.2em}
  \scalebox{0.63}{
  \begin{tabular}{@{\extracolsep{4pt}}lccccccc@{}}
    \toprule
    & \multicolumn{4}{c}{\textbf{SketchBoost (GPU)}} & \multicolumn{3}{c}{\textbf{Baseline (CPU/GPU)}} 
    \vspace{0.4em}\\
    \cline{2-5}  \cline{6-8} \vspace{-0.2em}
    \textbf{Dataset} & \textbf{Top Outputs} & \textbf{Random Sampling} & \textbf{Random Projection} & \textbf{SketchBoost Full} & \textbf{CatBoost} & \textbf{XGBoost} & \textbf{TabNet} \\
     &\small{(for the best $k$)} & \small{(for the best $k$)} & \small{(for the best $k$)} & \small{(multioutput)} & \small{(multioutput)} & \small{(one-vs-all)} & \small{(multioutput)}\\
    \midrule
    \textbf{Multiclass classification  } &  & & &  & \textbf{GPU} &\textbf{GPU} &\textbf{GPU} \\
    Otto (9 classes)& 113 & 102 & 89 & 131 & \textbf{73} & 1244 & 903 \\
    SF-Crime (39 classes) & 705 & 676 & \textbf{612} & 1146 & 659 & 4016 & 2683\\
    Helena (100 classes) & 154 & 180 & \textbf{113} & 355 & 436 & 1036 & 1196\\
    Dionis (355 classes) & 1889 & 2038 & \textbf{419} & 23919 & 18600 & 18635 & 1853\\
    \midrule
    \textbf{Multilabel classification} &  & & & &\textbf{CPU}  & \textbf{GPU} &\textbf{GPU}\\
    Mediamill (101 labels) & \textbf{251} & 263 & 294 & 1777 & 10164  & 2074 & 1231\\
    MoA (206 labels) & 103 & 189 & \textbf{87} & 696 & 9398 & 376 & 672\\
    Delicious (983 labels) & \textbf{575} & 664 & 1259 & 19553 & 20120 & 15795 & 2902\\
    \midrule
    \textbf{Multitask regression} &  & & & &\textbf{CPU}  & \textbf{GPU} &\textbf{GPU}\\
    RF1 (8 tasks) & 369 & 396 & 340 & 413 & 804 & {315} & \textbf{207}\\
    SCM20D (16 tasks) & 499 & 528 & {479} & 597 & 798 & 1432 & \textbf{296} \\
    \bottomrule
  \end{tabular}}
  \vspace{-1em}
\end{table}

\par
\vspace{-0.5em}
\paragraph{Results.} 
The final test errors are summarized in \Cref{tb:test_score}. 
Experiments show that, in general, SketchBoost with a sketching strategy obtains results comparable to or even better than the competing boosting frameworks.
Promisingly, there is always a sketching strategy that outperforms SketchBoost Full. 
Random Projection achieves the best scores, but Random Sampling also performs quite well. The deterministic Top Outputs strategy scores less than other baselines everywhere. In addition, it is noticeable that the one-vs-all strategy implemented in XGBoost leads to a worse generalization ability than the single-tree strategy on most datasets.
\par
\begin{figure}[t!]
\captionsetup{justification=centering}
\centering
\includegraphics[width=0.245\linewidth]{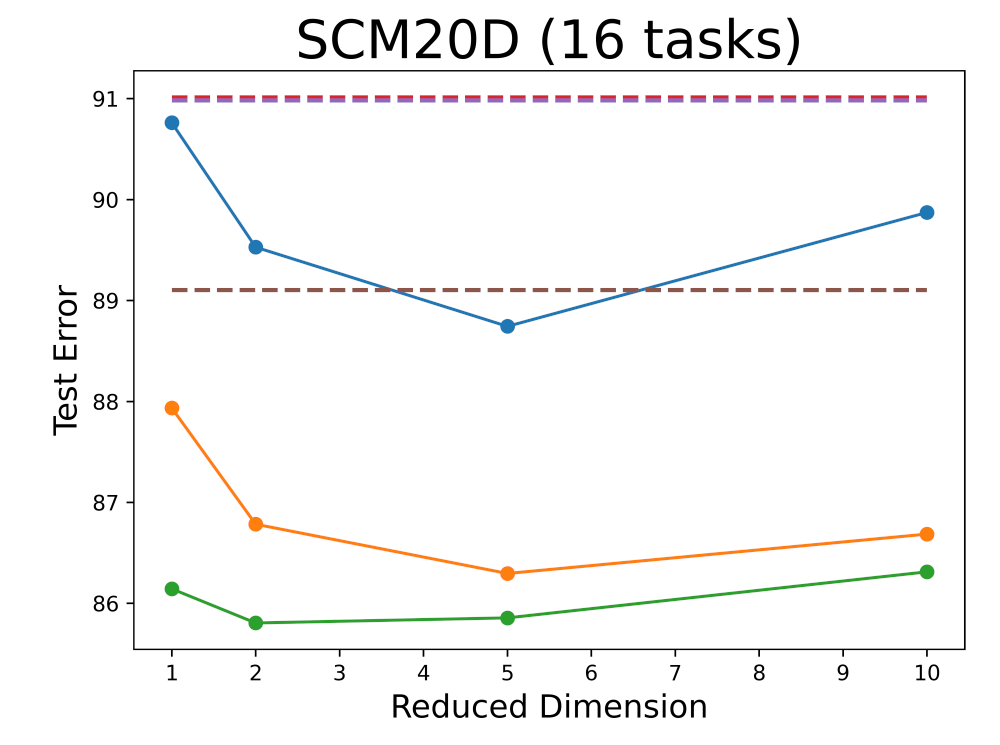}
\includegraphics[width=0.245\linewidth]{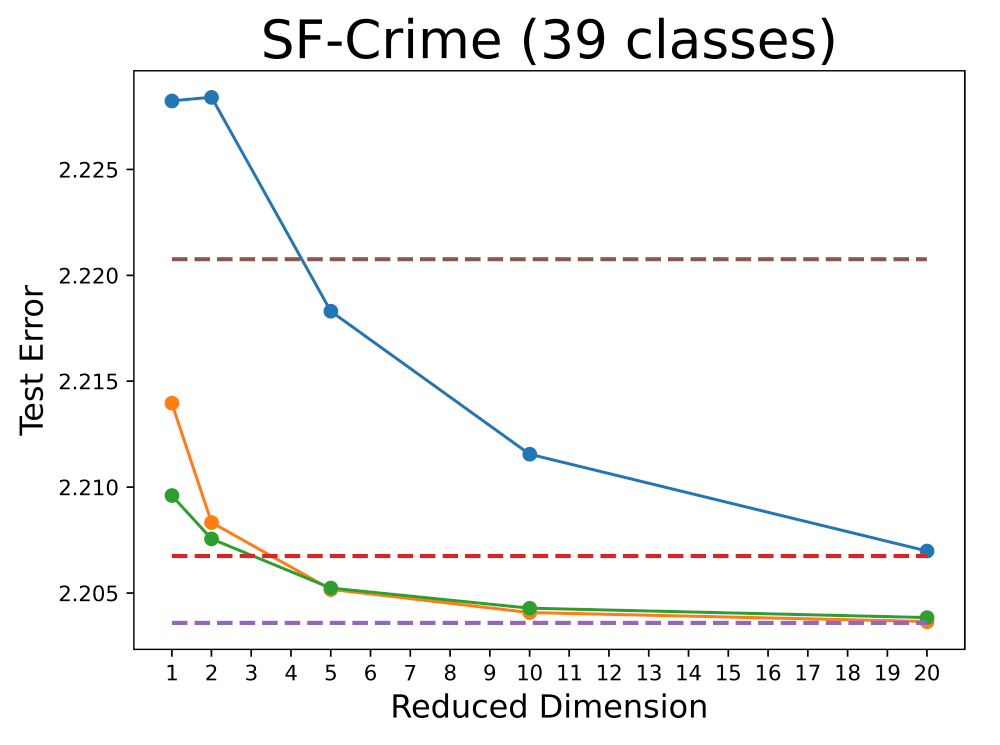}
\includegraphics[width=0.245\linewidth]{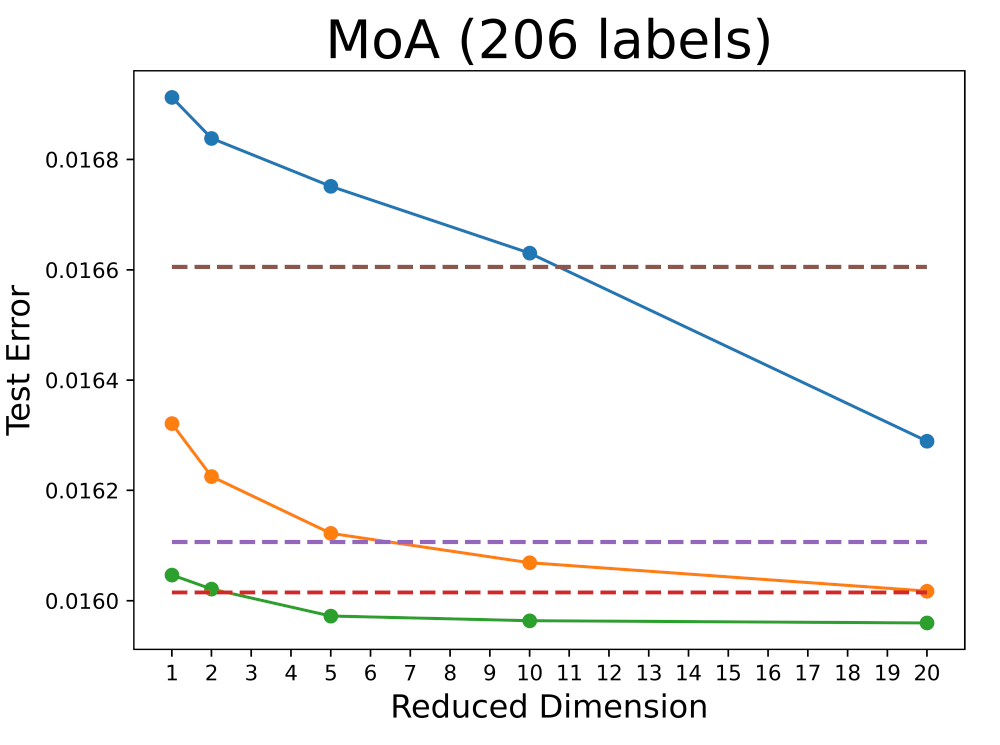}
\includegraphics[width=0.245\linewidth]{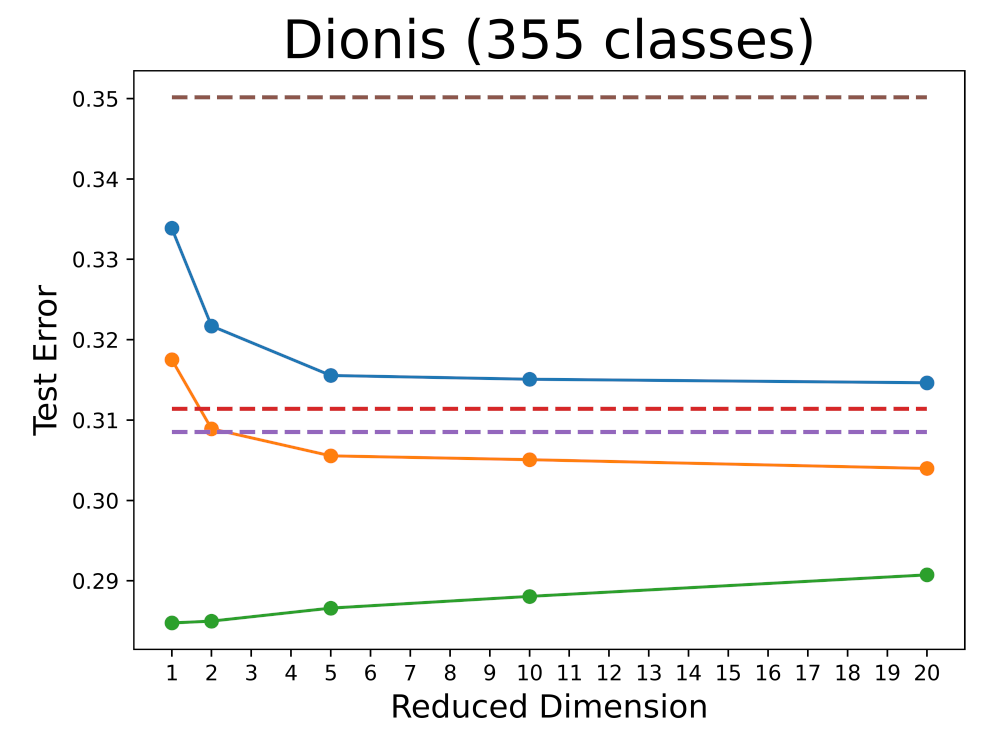}
\\
\includegraphics[width=0.95\linewidth]{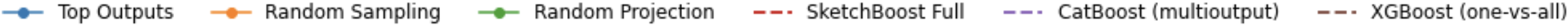}
\caption{\small Dependence of test errors (cross-entropy for classification and RMSE for regression) \\ on sketch dimension $\rdim$. \label{pic:test_score}}
\vspace{-1em}
\end{figure}

The dependence of test scores on the sketch size $\rdim$ for 
four datasets is shown in \Cref{pic:test_score}; 
for other datasets see \Cref{pic:test_score_appendix} in \Cref{sec:experiment_details}. 
It confirms the idea that, in general, the larger values $\rdim$ we take, the better performance we obtain. 
Moreover, our numerical study shows that there is a wide range of values of $\rdim$ for which sketching strategies work well; 
see the detailed results for all $\rdim$ 
in \Cref{tb:detailed_test_score_bce} and \Cref{tb:detailed_test_score_acc} in \Cref{sec:experiment_details}.
For most datasets, $\rdim\leq10$ is enough to obtain a result similar to or even better than SketchBoost Full or other baselines.
Loosely speaking, an intuitive explanation of why reducing the output dimension may increase the ensemble quality is that building a tree using all outputs often leads to bad split choices for some particular outputs. Sketching strategies use small groups of outputs, which leads to better tree structures for these outputs and a more diverse ensemble overall. In this connection, the optimal value of $\rdim$ strongly depends on the relations between the outputs in a given dataset.
With limited resources in practice, we would recommend using a predefined value $\rdim=5$. It is common in GBDTs: modern toolkits have more than $100$ hyperparameters, and many of them are not usually tuned (default values typically work well). 
But at the same time, one can always add $\rdim$ to the set of hyperparameters that are tuned. In our view, an additional hyperparameter will not play a significant role here taking into account that hyperparameter optimization is usually done using the random search or Bayesian optimization.
\par
\begin{figure}[t!]
\captionsetup{justification=centering}
\centering
\includegraphics[width=0.27\linewidth]{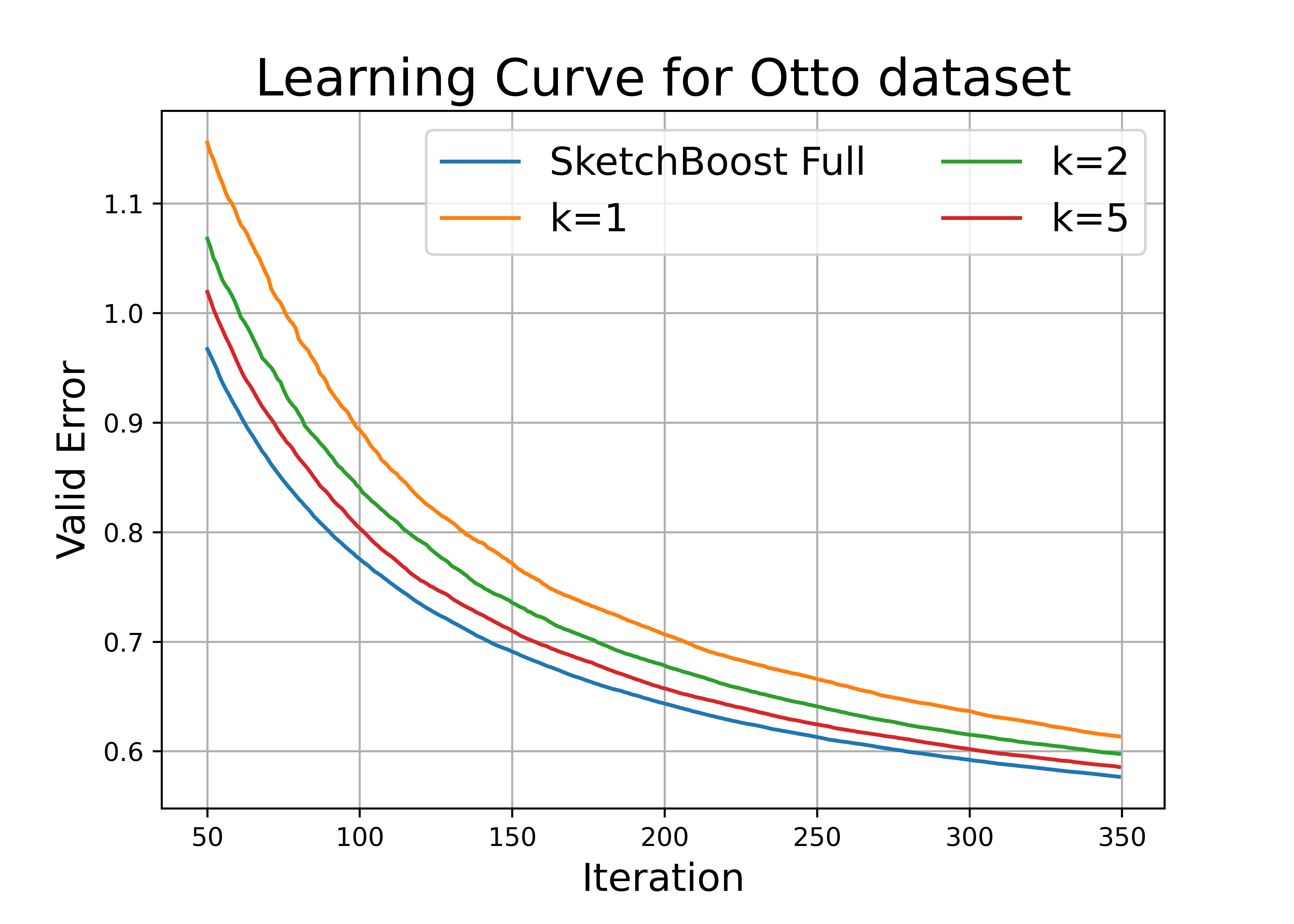}
\includegraphics[width=0.27\linewidth]{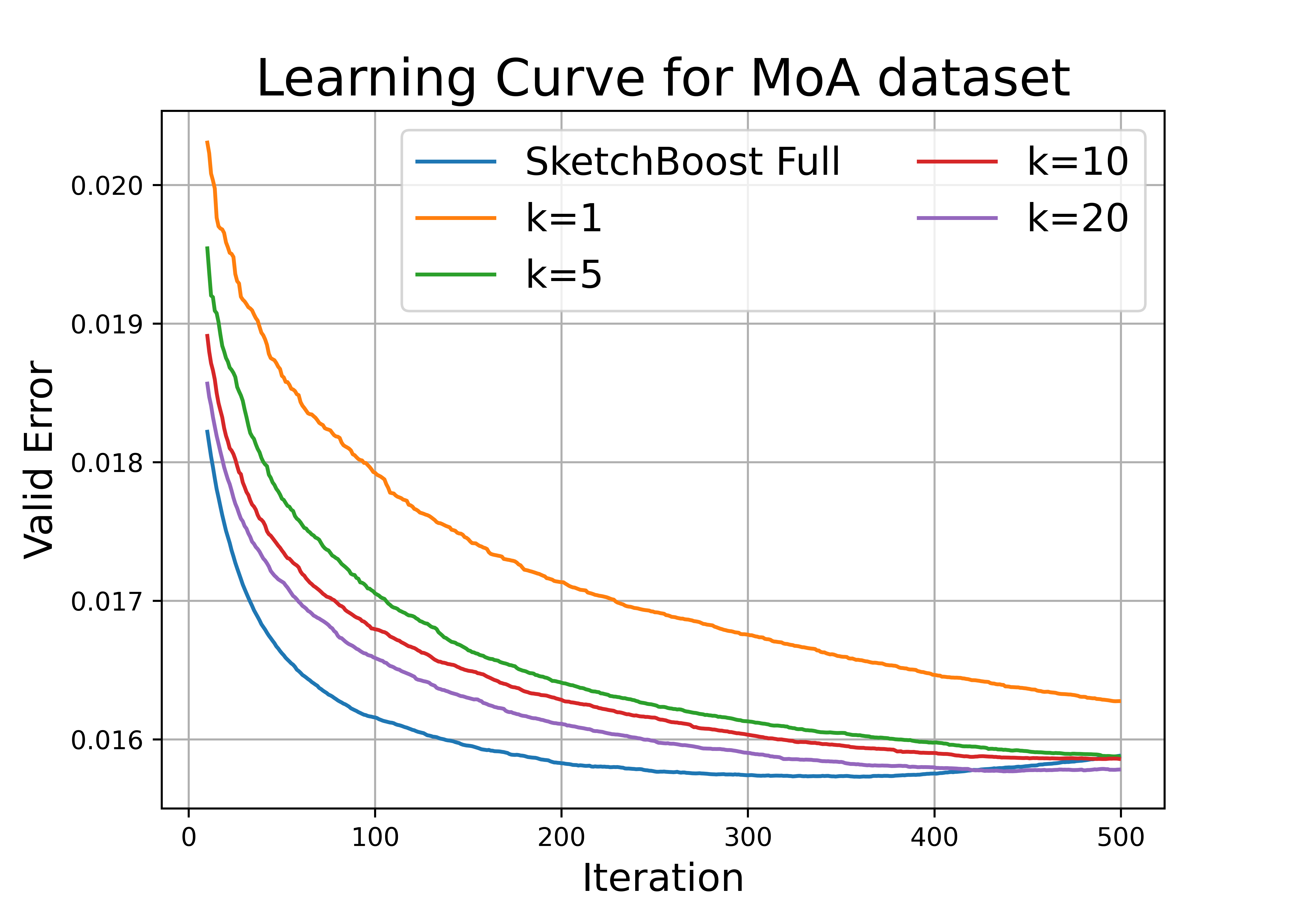}
\\
\caption{\small Learning curves for validation error for SketchBoost Full and SketchBoost with Random Sampling.\label{pic:learning_curve}}
\vspace{-1em}
\end{figure}

Further, the learning curves for validation errors on some datasets are given in \Cref{pic:learning_curve}. In general, it shows that small values of $\rdim$ result in a slower error decay at early iterations. But if $\rdim$ is properly defined, the validation error of SketchBoost with a sketching strategy is comparable to the error of SketchBoost Full, and hence both algorithms need approximately the same number of steps to convergence. This means that the proposed sketching strategies do not result in more complex models and do not significantly affect the model size or inference time. 
Detailed information on the number of steps to convergence for all strategies and baselines is given in \Cref{tb:best_iter} in \Cref{sec:experiment_details}.
\par
SketchBoost does a good job in reducing the training time. 
In \Cref{tb:train_time} we compare training times for SketchBoost, XGBoost, CatBoost, and TabNet. 
One can see that it 
significantly increases with the dataset size and, 
in particular, the output dimension. 
If a dataset is small,
as, for example, RF1 ($8$~targets, $9$k~rows, $64$ features) or Otto ($9$~classes, $61$k~rows, $93$ features), our Python implementation is slightly slower than the efficient CatBoost or XGBoost GPU implementations written on low-level programming languages. But for Dionis ($355$~classes,  $416$k~rows, $60$ features), our implementation together with a sketching strategy becomes $40$ times faster than XGBoost or CatBoost without sacrificing performance. 
Overall, we can conclude that
the proposed sketching algorithms can 
significantly speed up SketchBoost Full  
and can lead to considerably faster training than other GBDT baselines.
We recall that CatBoost can be trained on GPU only for multiclass classification tasks, and hence the time comparison with other algorithms on other tasks is not fair for CatBoost.
\par
Finally, 
we see that
all the GBDT implementations outperform TabNet 
in terms of test score on almost all tasks; 
see \Cref{tb:test_score} again. These results confirm the conclusion from the recent surveys \citep*{borisov2021deep} and \citep*{qin2021neural} that algorithms based on gradient-boosted tree ensembles still mostly outperform deep learning models on tabular supervised learning tasks. 
Nevertheless, \Cref{tb:train_time} shows that TabNet converges faster than GBDTs without sketching strategies. Moreover, TabNet is even faster than SketchBoost with sketching strategies on two regression tasks. The reason for this is that if the target dimension is high, it affects the complexity of a neural network only in the last layer and, in general, has little effect on the training time.  Further, it is also worth mentioning that neural networks tend to have much more hyperparameters than GBDTs and, as the result, need more time to be properly fine-tuned. 
Further details on this experiment are given in \Cref{sec:experiment_design_tabnet}.

\par
\par
\par
\vspace{-0.5em}
\paragraph{Comparison with GBDT-MO.} 
We also compare SketchBoost with GBDT-MO Full and GBDT-MO (sparse) from \cite{gbdt-mo-2021} (we want to highlight that GBDT-Sparse from \cite{gbdt-sparse-2017} does not have an open-source implementation). 
As sketching strategies, we consider here only Random Sampling and Random Projection. As the baseline, we consider only CatBoost on CPU
(to make it comparable to GBDT-MO which works only on CPU). The datasets to compare and the best hyperparameters were taken from the original paper. 
\par
Summary results are presented in \Cref{tb:test_score_gbdtmo} and \Cref{tb:train_time_gbdtmo}. SketchBoost with sketching strategies outperforms other algorithms on most datasets in terms of accuracy. GBDT-MO (sparse) is everywhere slower than GBDT-MO Full (because of optimization with a sparsity constraint).
Furthermore, its training time is comparable to CatBoost.
The time comparison with SketchBoost is not fair because of the GPU training, but, as it is shown, it is orders of magnitude faster. It is worth noting that SketchBoost Full is sometimes faster than SketchBoost with a sketching strategy. 
The reason for this is that if the dataset is small, then each boosting iteration requires little time. Therefore, when a sketching strategy is used, the speed up for each boosting iteration may be insignificant (especially because of ineffective utilization of GPU). At the same time, the number of iterations needed to convergence may be greater, which may result in an increase in the overall training time. Exactly this happened here. 
Further details on this experiment are given in \Cref{sec:experiment_gbdtmo}. 
\par
\begin{table}[t!]
  \setlength\tabcolsep{3pt}
  \setlength\extrarowheight{2pt}
  \captionsetup{justification=centering}
  \centering
  \vspace{-1em}
  \caption{\small
  Test scores (accuracy for classification and RMSE for regression)
  $\pm$ their standard deviation.\label{tb:test_score_gbdtmo}}
  \vspace{0.2em}
  \scalebox{0.63}{
  \begin{tabular}{@{\extracolsep{4pt}}lcccccc@{}}
    \toprule
    & \multicolumn{3}{c}{\textbf{SketchBoost}} & \multicolumn{2}{c}{\textbf{GBDT-MO}} 
    & \textbf{Baseline}
    \vspace{0.4em}\\
    \cline{2-4}  \cline{5-6} \cline{7-7} \vspace{-0.2em}
    \textbf{Dataset} & \textbf{Random Sampling} & \textbf{Random Projection} & \textbf{SketchBoost Full} & \textbf{GBDT-MO (sparse)} & \textbf{GBDT-MO Full} & \textbf{CatBoost} \\
     &\small{(for the best $k$)} & \small{(for the best $k$)} & \small{(multioutput)} & \small{(for the best $k$)} & \small{(multioutput)} & \small{(multioutput)}\\
    \midrule
    \textbf{Multiclass classification} & & & & & &\\
    MNIST (10 classes) & 0.9755 & 0.9740 & 0.9730 &  0.9758 & \textbf{0.9760} & 0.9684\\
    & \qquad\tcgr{$\pm$0.0042} & \qquad\tcgr{$\pm$0.0032} & \qquad\tcgr{$\pm$0.0028} &  \qquad\tcgr{$\pm$0.0048} & \qquad\tcgr{$\pm$0.0040} & \qquad\tcgr{$\pm$0.0040}\\
    Caltech (101 classes) & \textbf{0.5704} & 0.5623 & 0.5549 & 0.4796 & 0.4469 & 0.5049\\
    & \qquad\tcgr{$\pm$0.0273} & \qquad\tcgr{$\pm$0.0159} & \qquad\tcgr{$\pm$0.0080} & \qquad\tcgr{$\pm$0.0375} & \qquad\tcgr{$\pm$0.0590} & \qquad\tcgr{$\pm$0.0167}\\
    \midrule
    \textbf{Multilabel classification} & & & & & &\\
    NUS-WIDE (81 labels) & 0.9892 & \textbf{0.9897} & 0.9893&  0.9892 & 0.9891 & 0.9893\\
    & \qquad\tcgr{$\pm$0.0003} & \qquad\tcgr{$\pm$0.0003} & \qquad\tcgr{$\pm$0.0002} & \qquad\tcgr{$\pm$0.0006} & \qquad\tcgr{$\pm$0.0002} & \qquad\tcgr{$\pm$0.0001}\\
    \midrule
    \textbf{Multitask regression} & & & & & &\\
    MNIST-REG (24 tasks) & 0.2661 & \textbf{0.2654} & 0.2660 & 0.2736 & 0.2723 &  0.2708\\
    & \qquad\tcgr{$\pm$0.0019} & \qquad\tcgr{$\pm$0.0012} & \qquad\tcgr{$\pm$0.0019} & \qquad\tcgr{$\pm$0.0017} & \qquad\tcgr{$\pm$0.0026} & \qquad\tcgr{$\pm$0.0023}\\
    \bottomrule
  \end{tabular}}
\end{table}

\begin{table}[t!]
  \setlength\tabcolsep{3pt}
  \setlength\extrarowheight{2pt}
  \captionsetup{justification=centering}
  \centering
  \vspace{-1em}
  \caption{\small
  Training time per fold in seconds.\label{tb:train_time_gbdtmo}}
  \vspace{0.2em}
  \scalebox{0.63}{
  \begin{tabular}{@{\extracolsep{4pt}}lcccccc@{}}
    \toprule
    & \multicolumn{3}{c}{\textbf{SketchBoost (GPU)}} & \multicolumn{2}{c}{\textbf{GBDT-MO (CPU)}} 
    & \textbf{Baseline (CPU)}
    \vspace{0.4em}\\
    \cline{2-4}  \cline{5-6} \cline{7-7}  \vspace{-0.2em}
    \textbf{Dataset} & \textbf{Random Sampling} & \textbf{Random Projection} & \textbf{SketchBoost Full} & \textbf{GBDT-MO (sparse)} & \textbf{GBDT-MO Full} & \textbf{CatBoost} \\
     &\small{(for the best $k$)} & \small{(for the best $k$)} & \small{(multioutput)} & \small{(for the best $k$)} & \small{(multioutput)} & \small{(multioutput)}\\
    \midrule
    \textbf{Multiclass classification} & & & & & &\\
    MNIST (10 classes) & 102 & 66 & \textbf{46} & 399 & 362 & 156\\
    Caltech (101 classes) & 15 & 16 & \textbf{13} & 1312 & 776 & 136\\
    \midrule
    \textbf{Multilabel classification} & & & & & &\\
    NUS-WIDE (81 labels) & \textbf{36} & 72 & 87 & 3660 & 2606 & 13857\\
    \midrule
    \textbf{Multitask regression} & & & & & &\\
    MNIST-REG (24 tasks) & 120 & \textbf{45} & 90 & 163 & 210 & 964\\
    \bottomrule
  \end{tabular}}
  \vspace{-1em}
\end{table}
\par
\vspace{-0.5em}
\section{Conclusion}
\label{sec:conclusion}

\begin{wrapfigure}{r}{0.47\textwidth}
\vspace{-2.5em}
\centering
    \includegraphics[width=\linewidth]{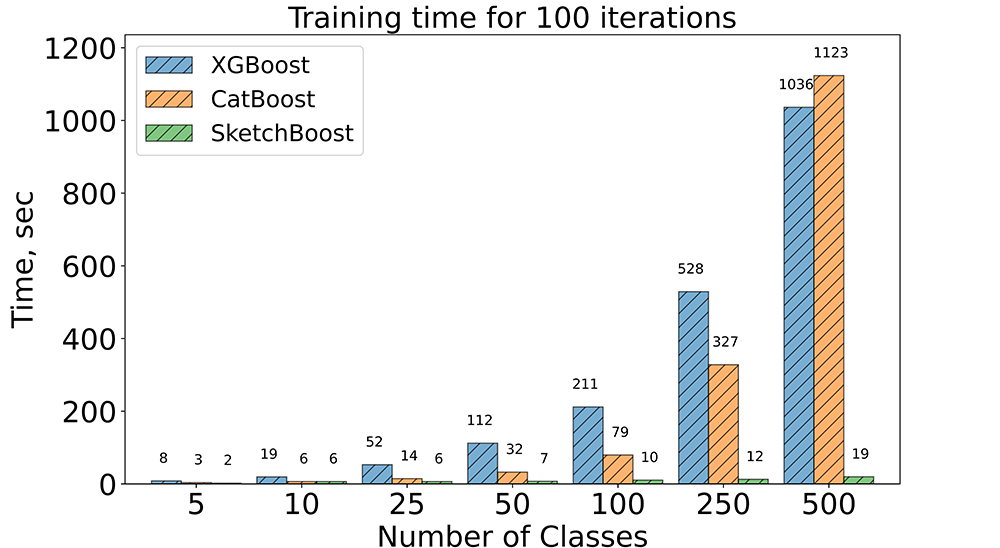}
    \caption{\small Training time of XGBoost, CatBoost, and SkechBoost in the same experiment as in \Cref{pic:train_time_main}. Here SketchBoost 
    uses Random Projection with sketch dimension $\rdim=5$. Further details are given in \Cref{sec:synthetic_dataset}.\label{pic:train_time_main_SB}}
\vspace{-1em}
\end{wrapfigure}
\par
In this paper, we presented effective 
methods to speed up GBDT on multioutput tasks.
These methods are generic and can be easily
integrated into any single-tree GBDT realization.
On real-world datasets, these methods achieve 
comparable and sometimes even better results
to the existing state-of-the-art 
GBDT implementations but in remarkably less time.
The proposed methods are implemented in SketchBoost
which itself is a part of 
our Python-based implementation of GBDT
called Py-Boost. 
\Cref{pic:train_time_main_SB} concludes this paper 
by showing the gain in training time of SkechBoost 
in the same experiment as in \Cref{pic:train_time_main} 
from the Introduction.

\section*{Acknowledgements}
\label{sec:acknowledgements}
We would like to thank 
Gleb Gusev and Bulat Ibragimov for
helpful discussions and feedback for an 
earlier draft of this work,
Dmitry Simakov and Mikhail Kuznetsov 
for the help with the TabNet experiments, 
and Maxim Savchenko and all the Sber AI Lab team for
their support and active interest in this project.
We would also like to thank the anonymous reviewers for their thoughtful feedback.

\clearpage
\bibliographystyle{plainnat}
\bibliography{bibliography}

\begin{thebibliography}{49}
\providecommand{\natexlab}[1]{#1}
\providecommand{\url}[1]{\texttt{#1}}
\expandafter\ifx\csname urlstyle\endcsname\relax
  \providecommand{\doi}[1]{doi: #1}\else
  \providecommand{\doi}{doi: \begingroup \urlstyle{rm}\Url}\fi

\bibitem[Achlioptas(2003)]{achlioptas-2003}
Dimitris Achlioptas.
\newblock {Database-friendly random projections: Johnson-Lindenstrauss with
  binary coins}.
\newblock \emph{Journal of Computer and System Sciences}, 66\penalty0
  (4):\penalty0 671--687, 2003.

\bibitem[Ailon and Chazelle(2009)]{ailon-chazelle-2009}
Nir Ailon and Bernard Chazelle.
\newblock {The Fast Johnson--Lindenstrauss Transform and Approximate Nearest
  Neighbors}.
\newblock \emph{SIAM Journal on Computing}, 39:\penalty0 302--322, 2009.

\bibitem[Akiba et~al.(2019)Akiba, Sano, Yanase, Ohta, and
  Koyama]{akiba2019optuna}
Takuya Akiba, Shotaro Sano, Toshihiko Yanase, Takeru Ohta, and Masanori Koyama.
\newblock {Optuna: A Next-Generation Hyperparameter Optimization Framework}.
\newblock In \emph{Proceedings of the 25th ACM SIGKDD international conference
  on knowledge discovery \& data mining}, KDD '19, pages 2623--2631, 2019.

\bibitem[Alsabti et~al.(1998)Alsabti, Ranka, and Singh]{ars-1998}
Khaled Alsabti, Sanjay Ranka, and Vineet Singh.
\newblock {CLOUDS: A Decision Tree Classifier for Large Datasets}.
\newblock In \emph{Proceedings of the Fourth International Conference on
  Knowledge Discovery and Data Mining}, KDD'98, pages 2--8. AAAI Press, 1998.

\bibitem[Appel et~al.(2013)Appel, Fuchs, Doll\'{a}r, and Perona]{afdp-2013}
Ron Appel, Thomas Fuchs, Piotr Doll\'{a}r, and Pietro Perona.
\newblock {Quickly Boosting Decision Trees: Pruning Underachieving Features
  Early}.
\newblock In \emph{Proceedings of the 30th International Conference on
  International Conference on Machine Learning}, number~3 in ICML'13, pages
  594--602. PMLR, 2013.

\bibitem[Arik and Pfister(2021)]{arik2021tabnet}
Sercan~{\"O}. Arik and Tomas Pfister.
\newblock {Tabnet: Attentive Interpretable Tabular Learning}.
\newblock \emph{Proceedings of the AAAI Conference on Artificial Intelligence},
  35\penalty0 (8):\penalty0 6679--6687, 2021.

\bibitem[Bergstra et~al.(2011)Bergstra, Bardenet, Bengio, and
  K\'{e}gl]{bergstra2011algorithms}
James Bergstra, R\'{e}mi Bardenet, Yoshua Bengio, and Bal\'{a}zs K\'{e}gl.
\newblock {Algorithms for Hyper-Parameter Optimization}.
\newblock In \emph{Proceedings of the 24th International Conference on Neural
  Information Processing Systems}, NIPS'11, pages 2546--2554. Curran Associates
  Inc., 2011.

\bibitem[Borisov et~al.(2021)Borisov, Leemann, Se{\ss}ler, Haug, Pawelczyk, and
  Kasneci]{borisov2021deep}
Vadim Borisov, Tobias Leemann, Kathrin Se{\ss}ler, Johannes Haug, Martin
  Pawelczyk, and Gjergji Kasneci.
\newblock {Deep Neural Networks and Tabular Data: A Survey}, 2021.

\bibitem[Breiman(1996)]{breiman1996bagging}
Leo Breiman.
\newblock {Bagging Predictors}.
\newblock \emph{Machine learning}, 24\penalty0 (2):\penalty0 123--140, 1996.

\bibitem[Chen and Guestrin(2016)]{xgboost-2016}
Tianqi Chen and Carlos Guestrin.
\newblock {XGBoost: A Scalable Tree Boosting System}.
\newblock In \emph{Proceedings of the 22nd ACM SIGKDD International Conference
  on Knowledge Discovery and Data Mining}, KDD '16, pages 785--794. Association
  for Computing Machinery, 2016.

\bibitem[Ciss\'{e} et~al.(2012)Ciss\'{e}, Arti\`{e}res, and Gallinari]{cag2012}
M.~Ciss\'{e}, T.~Arti\`{e}res, and Patrick Gallinari.
\newblock {Learning Compact Class Codes for Fast Inference in Large Multi Class
  Classification}.
\newblock In \emph{Proceedings of the 2012th European Conference on Machine
  Learning and Knowledge Discovery in Databases - Volume Part I}, ECMLPKDD'12,
  pages 506--520, 2012.

\bibitem[Dasgupta et~al.(2010)Dasgupta, Kumar, and Sarlos]{dasgupta-2010}
Anirban Dasgupta, Ravi Kumar, and Tam\'{a}s Sarlos.
\newblock {A Sparse Johnson: Lindenstrauss Transform}.
\newblock In \emph{Proceedings of the Forty-Second ACM Symposium on Theory of
  Computing}, STOC '10, pages 341--350. Association for Computing Machinery,
  2010.

\bibitem[Fiedler(2021)]{fiedler2021simple}
James Fiedler.
\newblock {Simple Modifications to Improve Tabular Neural Networks}, 2021.

\bibitem[Freund(1995)]{freund-1995}
Yoav Freund.
\newblock {B}oosting a {W}eak {L}earning {A}lgorithm by {M}ajority.
\newblock \emph{Information and Computation}, 121\penalty0 (2):\penalty0
  256--285, 1995.

\bibitem[Freund and Schapire(1997)]{freund-schapire-1997}
Yoav Freund and Robert~E Schapire.
\newblock {A} {D}ecision-{T}heoretic {G}eneralization of {O}n-{L}ine {L}earning
  and an {A}pplication to {B}oosting.
\newblock \emph{Journal of Computer and System Sciences}, 55\penalty0
  (1):\penalty0 119--139, 1997.

\bibitem[Friedman(2001)]{friedman-2001}
Jerome~H. Friedman.
\newblock {Greedy function approximation: {A} gradient boosting machine.}
\newblock \emph{The Annals of Statistics}, 29\penalty0 (5):\penalty0 1189 --
  1232, 2001.

\bibitem[Friedman(2002)]{friedman-2002}
Jerome~H. Friedman.
\newblock {Stochastic Gradient Boosting}.
\newblock \emph{Computational Statistics \& Data Analysis}, 38\penalty0
  (4):\penalty0 367--378, 2002.

\bibitem[Golub and Van~Loan(1996)]{golub-1996}
Gene~H. Golub and Charles~F. Van~Loan.
\newblock \emph{{Matrix Computations}}.
\newblock The Johns Hopkins University Press, third edition, 1996.

\bibitem[Gorishniy et~al.(2021)Gorishniy, Rubachev, Khrulkov, and
  Babenko]{gorishniy2021revisiting}
Yury Gorishniy, Ivan Rubachev, Valentin Khrulkov, and Artem Babenko.
\newblock {Revisiting Deep Learning Models for Tabular Data}.
\newblock \emph{Advances in Neural Information Processing Systems},
  34:\penalty0 18932--18943, 2021.

\bibitem[Guyon(2003)]{guyon2003design}
Isabelle Guyon.
\newblock {Design of experiments of the NIPS 2003 variable selection
  benchmark}.
\newblock In \emph{NIPS 2003 workshop on feature extraction and feature
  selection}, volume 253, page~40, 2003.

\bibitem[Holodnak and Ipsen(2015)]{holodnak-ipsen-2015}
John~T. Holodnak and Ilse C.~F. Ipsen.
\newblock {Randomized Approximation of the Gram Matrix: Exact Computation and
  Probabilistic Bounds}.
\newblock \emph{SIAM Journal on Matrix Analysis and Applications}, 36:\penalty0
  110--137, 2015.

\bibitem[Hsu et~al.(2009)Hsu, Kakade, Langford, and Zhang]{hklz2009}
Daniel Hsu, Sham~M. Kakade, John Langford, and Tong Zhang.
\newblock {Multi-Label Prediction via Compressed Sensing}.
\newblock In \emph{Proceedings of the 22nd International Conference on Neural
  Information Processing Systems}, NIPS'09, pages 772--780. Curran Associates
  Inc., 2009.

\bibitem[Ibragimov and Gusev(2019)]{mvs-2019}
Bulat Ibragimov and Gleb Gusev.
\newblock {Minimal {V}ariance Sampling in Stochastic Gradient Boosting}, 2019.

\bibitem[Indyk and Motwani(1998)]{indyk-motwani-1998}
Piotr Indyk and Rajeev Motwani.
\newblock {Approximate Nearest Neighbors: Towards Removing the Curse of
  Dimensionality}.
\newblock In \emph{Proceedings of the Thirtieth Annual ACM Symposium on Theory
  of Computing}, STOC '98, pages 604--613, 1998.

\bibitem[Jahrer et~al.(2010)Jahrer, T\"{o}scher, and Legenstein]{jtl-2010}
Michael Jahrer, Andreas T\"{o}scher, and Robert Legenstein.
\newblock {Combining Predictions for Accurate Recommender Systems}.
\newblock In \emph{Proceedings of the 16th ACM SIGKDD International Conference
  on Knowledge Discovery and Data Mining}, KDD '10, pages 693--702, New York,
  NY, USA, 2010. Association for Computing Machinery.

\bibitem[Jimenez and Landgrebe(1999)]{jl-1999}
L.O. Jimenez and D.A. Landgrebe.
\newblock {Hyperspectral Data Analysis and Supervised Feature Reduction via
  Projection Pursuit}.
\newblock \emph{IEEE Transactions on Geoscience and Remote Sensing},
  37\penalty0 (6):\penalty0 2653--2667, 1999.

\bibitem[Jin and Agrawal(2003{\natexlab{a}})]{jin-agrawal-2003}
Ruoming Jin and Gagan Agrawal.
\newblock {Efficient Decision Tree Construction on Streaming Data}.
\newblock In \emph{Proceedings of the Ninth ACM SIGKDD International Conference
  on Knowledge Discovery and Data Mining}, KDD'03, pages 571--576, New York,
  NY, USA, 2003{\natexlab{a}}. Association for Computing Machinery.

\bibitem[Jin and Agrawal(2003{\natexlab{b}})]{rg-2003}
Ruoming Jin and Gagan Agrawal.
\newblock {Communication and Memory Efficient Parallel Decision Tree
  Construction}.
\newblock In \emph{Proceedings of the Third {SIAM} International Conference on
  Data Mining}, pages 119--129. {SIAM}, 2003{\natexlab{b}}.

\bibitem[Johnson and Lindenstrauss(1984)]{johnson-lindenstrauss-1984}
William Johnson and Joram Lindenstrauss.
\newblock Extensions of lipschitz maps into a hilbert space.
\newblock \emph{Contemporary Mathematics}, 26:\penalty0 189--206, 1984.

\bibitem[Kane and Nelson(2014)]{kane-nelson-2014}
Daniel~M. Kane and Jelani Nelson.
\newblock {Sparser Johnson-Lindenstrauss Transforms}.
\newblock \emph{Journal of the ACM}, 61\penalty0 (1), 2014.

\bibitem[Kapoor et~al.(2012)Kapoor, Viswanathan, and Jain]{kvj2012}
Ashish Kapoor, Raajay Viswanathan, and Prateek Jain.
\newblock {Multilabel Classification using Bayesian Compressed Sensing}.
\newblock In F.~Pereira, C.J. Burges, L.~Bottou, and K.Q. Weinberger, editors,
  \emph{Advances in Neural Information Processing Systems}, volume~25 of
  \emph{NIPS'12}, pages 2645--2653. Curran Associates, Inc., 2012.

\bibitem[Ke et~al.(2017)Ke, Meng, Finley, Wang, Chen, Ma, Ye, and
  Liu]{lightgbm-2017}
Guolin Ke, Qi~Meng, Thomas Finley, Taifeng Wang, Wei Chen, Weidong Ma, Qiwei
  Ye, and Tie-Yan Liu.
\newblock {LightGBM: A Highly Efficient Gradient Boosting Decision Tree}.
\newblock In \emph{Proceedings of the 31st International Conference on Neural
  Information Processing Systems}, NIPS'17, pages 3149--3157. Curran Associates
  Inc., 2017.

\bibitem[Kyrillidis et~al.(2014)Kyrillidis, Vlachos, and
  Zouzias]{kyrillidis-2014}
Anastasios Kyrillidis, Michail Vlachos, and Anastasios Zouzias.
\newblock {Approximate Matrix Multiplication with Application to Linear
  Embeddings}.
\newblock \emph{2014 IEEE International Symposium on Information Theory}, pages
  2182--2186, 2014.

\bibitem[Li et~al.(2007)Li, Burges, and Wu]{lbw-2007}
Ping Li, Christopher J.~C. Burges, and Qiang Wu.
\newblock {McRank: Learning to Rank Using Multiple Classification and Gradient
  Boosting}.
\newblock In \emph{Proceedings of the 20th International Conference on Neural
  Information Processing Systems}, NIPS'07, pages 897--904. Curran Associates
  Inc., 2007.

\bibitem[Li et~al.(2008)Li, Wu, and Burges]{lwb-2008}
Ping Li, Qiang Wu, and Christopher Burges.
\newblock {McRank: Learning to Rank Using Multiple Classification and Gradient
  Boosting}.
\newblock In \emph{Advances in Neural Information Processing Systems},
  volume~20. Curran Associates, Inc., 2008.

\bibitem[Mahoney(2011)]{mahoney-11}
Michael~W. Mahoney.
\newblock {Randomized Algorithms for Matrices and Data}.
\newblock \emph{Foundations and Trends in Machine Learning}, 3\penalty0
  (2):\penalty0 123--224, 2011.

\bibitem[Mehta et~al.(1996)Mehta, Agrawal, and Rissanen]{mrj-1996}
Manish Mehta, Rakesh Agrawal, and Jorma Rissanen.
\newblock {SLIQ: A Fast Scalable Classifier for Data Mining}.
\newblock In \emph{International Conference on Extending Database Technology},
  pages 18--32. Springer-Verlag, 1996.

\bibitem[Obermann and Waack(2016)]{oberman-waack-2016}
Lennart Obermann and Stephan Waack.
\newblock {Interpretable Multiclass Models for Corporate Credit Rating Capable
  of Expressing Doubt}.
\newblock \emph{Frontiers in Applied Mathematics and Statistics}, 2, 2016.

\bibitem[Prokhorenkova et~al.(2018)Prokhorenkova, Gusev, Vorobev, Dorogush, and
  Gulin]{catboost-2018}
Liudmila Prokhorenkova, Gleb Gusev, Aleksandr Vorobev, Anna~Veronika Dorogush,
  and Andrey Gulin.
\newblock {CatBoost: Unbiased Boosting with Categorical Features}.
\newblock In \emph{Proceedings of the 32nd International Conference on Neural
  Information Processing Systems}, NIPS'18, pages 6639--6649. Curran Associates
  Inc., 2018.

\bibitem[Qin et~al.(2021)Qin, Yan, Zhuang, Tay, Pasumarthi, Wang, Bendersky,
  and Najork]{qin2021neural}
Zhen Qin, Le~Yan, Honglei Zhuang, Yi~Tay, Rama~Kumar Pasumarthi, Xuanhui Wang,
  Mike Bendersky, and Marc Najork.
\newblock {Are Neural Rankers still Outperformed by Gradient Boosted Decision
  Trees?}
\newblock In \emph{International Conference on Learning Representations}, 2021.

\bibitem[Robert and Casella(2005)]{robert-casella-2005}
Christian~P. Robert and George Casella.
\newblock \emph{{Monte Carlo Statistical Methods (Springer Texts in
  Statistics)}}.
\newblock Springer-Verlag, Berlin, Heidelberg, 2005.
\newblock ISBN 0387212396.

\bibitem[Schapire(1990)]{shapire-1990}
Robert~E. Schapire.
\newblock {T}he {S}trength of {W}eak {L}earnability.
\newblock \emph{Machine Learning}, 5\penalty0 (2):\penalty0 197–227, 1990.

\bibitem[Si et~al.(2017)Si, Zhang, Keerthi, Mahajan, Dhillon, and
  Hsieh]{gbdt-sparse-2017}
Si~Si, Huan Zhang, S.~Sathiya Keerthi, Dhruv Mahajan, Inderjit~S. Dhillon, and
  Cho-Jui Hsieh.
\newblock Gradient {B}oosted {D}ecision {T}rees for {H}igh dimensional {S}parse
  {O}utput.
\newblock In \emph{Proceedings of the 34th International Conference on Machine
  Learning}, volume~70 of \emph{Proceedings of Machine Learning Research},
  pages 3182--3190. PMLR, 2017.

\bibitem[Tai and Lin(2012)]{tl2012}
Farbound Tai and Hsuan-Tien Lin.
\newblock {Multilabel Classification with Principal Label Space
  Transformation}.
\newblock \emph{Neural Computation}, 24\penalty0 (9):\penalty0 2508--2542,
  2012.

\bibitem[Wicker et~al.(2016)Wicker, Tyukin, and Kramer]{wtk2016}
J\"{o}rg Wicker, Andrey Tyukin, and Stefan Kramer.
\newblock {A Nonlinear Label Compression and Transformation Method for
  Multi-Label Classification using Autoencoders}.
\newblock In \emph{The 20th Pacific Asia Conference on Knowledge Discovery and
  Data Mining (PAKDD)}, volume 9651 of \emph{Lecture Notes in Computer
  Science}, pages 328--340, Switzerland, 2016. Springer International
  Publishing.

\bibitem[Woodruff(2014)]{wodruff-2014}
David~P. Woodruff.
\newblock {Sketching as a Tool for Numerical Linear Algebra}.
\newblock \emph{Foundations and Trends in Machine Learning}, 10\penalty0
  (1--2):\penalty0 1--157, 2014.

\bibitem[Zhai et~al.(2020)Zhai, Yao, and Zhou]{zyz-2020}
Naiju Zhai, Peifu Yao, and Xiaofeng Zhou.
\newblock {Multivariate Time Series Forecast in Industrial Process Based on
  XGBoost and GRU}.
\newblock In \emph{2020 IEEE 9th Joint International Information Technology and
  Artificial Intelligence Conference (ITAIC)}, volume~9, pages 1397--1400.
  IEEE, 2020.

\bibitem[Zhang and Jung(2021)]{gbdt-mo-2021}
Zhendong Zhang and Cheolkon Jung.
\newblock {GBDT-MO}: {G}radient-{B}oosted {D}ecision {T}rees for {M}ultiple
  {O}utputs.
\newblock \emph{IEEE Transactions on Neural Networks and Learning Systems},
  32:\penalty0 3156--3167, 2021.

\bibitem[Zhou(2012)]{zhou-2012}
Zhi-Hua Zhou.
\newblock \emph{{Ensemble Methods: Foundations and Algorithms}}.
\newblock Chapman and Hall/CRC, 2012.

\end{thebibliography}

\newpage
\appendix
\section{Additional Information on Sketched Split Scoring Methods.}
\label{sec:proofs}

This section provides additional information 
on sketched split scoring methods  from Section~3. 
Let us recall
that the scoring function $\Sf(R)$ for a leaf $R$ 
is given by
\[
    \Sf_{\G}(R) = \frac{\bigl\|\G^{\T}v_{R}\bigr\|^2}{|R|+ \lambda},
    \quad\text{where}\ \,
    \G =
    \begin{pmatrix}
    g_1^1 & g_1^2 & \ldots & g_1^{\odim}\\
    \vdots &  \vdots & \ddots & \vdots\\
    g_{\ssize}^1 & g_{\ssize}^2 & \ldots & g_{\ssize}^{\odim}\\
    \end{pmatrix}
    \ \,\text{and}\ \,
    v_{R} = 
    \begin{pmatrix}
    \I{x_1 \in R} \\
    \vdots \\
    \I{x_{\ssize} \in R} \\
    \end{pmatrix}.
\]
Here $\G\in\R^{\ssize \times \odim}$ is the matrix whose $i$-th row consists of  
gradient values 
$(g_i^1,\ldots,g_i^{\odim})= \nabla_a  \loss(y_i,a) 
|_{a=\alg_{t-1}(x_i)}$, $i=1,\ldots,\ssize$,
and $v_{R}$ is the indicator vector of leaf $R$ 
(its $i$-th component equals $1$ if $x_i\in R$ and $0$ otherwise). 
To reduce the complexity of computing $\Sf_{\G}(R)$ in $\odim$,
we approximate $\Sf_{\G}(R)$ with $\Sf_{\Gr}(R)$ 
for some sketch matrix $\Gr\in\R^{\ssize \times \rdim}$ 
with $\rdim \ll \odim$. 
The error of this approximation is measured by 
\begin{align}
    \Error(\Sf_{\G},\Sf_{\Gr}) 
    = \sup_{R}\bigl|\Sf_{\G}(R) - \Sf_{\Gr}(R)\bigr|.
    \label{eq:errorfunc2}
\end{align}
The supremum here is taken over all
possible leaves $R$, so that we aim at the optimization of the worst case. The reason for this is that 
we want the proposed approximation to be 
universal and uniformly accurate for all 
splits we will possibly iterate over. 
\par
Given the two matrices $\G$ and $\Gr$, 
the optimization problem from \eqref{eq:errorfunc2}
is an instance of Integer Programming problem 
and hence is NP-complete. To obtain a closed-form solution,
one needs to iterate through all possible leaves $R$ 
(that is, through all possible vectors $v_R$ with $0/1$ entries). Since the brute force is not an option 
in our case, we will replace this problem with 
a relaxed one and will look for nearly-optimal solutions.
We will show that reasonably good upper bounds 
on the error
are obtained when $\G\G^{\T}$ is well approximated
with $\Gr\Gr^{\T}$ in the operator norm; 
see \Cref{th:mainbound}.
This observation links
our problem to Approximate Matrix Multiplication (AMM).
In the next sections, we review some 
deterministic and random methods from AMM 
and apply them to construct nearly optimal sketches $\Gr$.

\paragraph{Auxiliary Lemma.}
First let us state an auxiliary lemma which 
bounds the approximation error
from \eqref{eq:errorfunc2} with
the distance between $\G \G^{\T}$ and $\Gr\Gr^{\T}$ 
is the operator norm.

\begin{lemma}
\label{th:mainbound}
Let $\G\in\R^{\ssize \times \odim}$ and 
$\Gr\in\R^{\ssize \times \rdim}$ be any two matrices.
Then 
\begin{align}
    \operatorname{Error}(\Sf_{\G},\Sf_{\Gr}) 
    \leq 
    \bigl\|\G \G^{\T} - \Gr\Gr^{\T}\bigr\|.
    \label{eq:mainbound}
\end{align}
\end{lemma}
\begin{proof}
A direct computation yields  
\begin{align*}
    \sup_{R}\bigl|\Sf_{\G}(R) - \Sf_{\Gr}(R)\bigr| 
    &= 
    \sup_{R}\left|\frac{\|\G^{\T}v_{R}\|^2 - \|\Gr^{\T}v_{R}\|^2}{|R|+ \lambda} \right|\\
    &\leq
    \sup_{R}\frac{\|\G\G^{\T} - \Gr\Gr^{\T}\| \|v_{R}\|^2}{|R|+ \lambda}.
\end{align*}
Since $\lambda>0$ and $\|v_{R}\|^2\leq|R|$ 
($v_{R}$ has $|R|$ non-zero entries equal to $1$), 
the assertion follows.
\end{proof}

Note that in practice we do not need to compute $\G \G^{\T}$. \Cref{th:mainbound} only provides a theoretical bound which will be used further. This bound is universal
for all possible leafs $R$ and involves only the gradient 
matrix $\G$ and its sketch $\Gr$.

\subsection{Truncated SVD}
We start with the Truncated SVD algorithm since,
by the matrix approximation lemma (the Eckart-Young-Mirsky theorem), 
it provides the optimal deterministic solution to AMM.
The following proposition summarizes its performance.

\begin{proposition} 
\label{th:svdbound}
Let $\G\in\R^{\ssize \times \odim}$ be any matrix.
Let also $\Gr\in\R^{\ssize \times \rdim}$ be the best 
$\rdim$-rank approximation of $\G$ 
provided by the Truncated SVD.
Then 
\begin{align*}
    \Error(\Sf_{\G},\Sf_{\Gr}) 
    \leq 
    \sigma^2_{\rdim+1}(\G),
\end{align*}
where $\sigma^2_{\rdim+1}(\G)$ is $(\rdim+1)$ largest  singular value of $\G$.
\end{proposition}
\begin{proof}
Let $\G = U \Sigma V^{\T}$ 
be the full SVD of $\G$.
Let also 
$\Gr = U_{\rdim} \Sigma_{\rdim}$ 
be the $\rdim$-rank Truncated SVD of $\G$ where we
keep only largest $\rdim$ singular values 
and corresponding columns in $U$. 
Using \Cref{th:mainbound}, we get 
\begin{align*}
    \sup_{R}\bigl|\Sf_{\G}(R) - \Sf_{\Gr}(R)\bigr| 
    \leq
    \bigl\|\G \G^{\T} - \Gr\Gr^{\T}\bigr\| 
    =
    \bigl\| U \Sigma^2 U^{\T} - U_{\rdim} \Sigma_{\rdim}^2 U_{\rdim}^{\T}\bigr\|. 
\end{align*}
Now the Eckart–Young–Mirsky theorem 
(for the spectral norm) yelds 
\begin{align*}
    \bigl\| U \Sigma^2 U^{\T} - U_{\rdim} \Sigma_{\rdim}^2 U_{\rdim}^{\T}\bigr\|
    = \sigma^2_{s+1}(\G),
\end{align*}
which finishes the proof.
\end{proof}

This proposition asserts that to speed up the 
split search, the gradient matrix $\G$
with $\odim$ columns can be replaced by its 
Truncated SVD estimate with $\rdim$ columns.
As a result, the scoring function $\Sf_{\G}$ will not change
significantly provided that $(\rdim+1)$ largest 
singular value of $\G$ is small.
The parameter $\rdim$ here can 
be chosen adaptively depending of the 
spectrum of $\G$ and values on $\Sf_{\G}$. 
\par
We haven't discussed Truncated SVD in the paper
due to its computational complexity which is 
$O(\min\{\ssize\odim^2,\ssize^2\odim\})$;
see \citep{golub-1996}. As it was discussed in
the Introduction, the computational complexity
of GBDT scales linearly in the output dimension $\odim$.
Consequently, the application of Truncated SVD
will only increase this complexity.
Further, we discuss methods 
with less computational costs.

\subsection{Top Outputs}

Top Outputs is a straightforward 
method which constructs the sketch $\Gr$ by 
keeping $\rdim$ columns of the 
gradient matrix $\G$ with the largest Euclidian norm.

\begin{proposition} 
\label{th:topkbound}
Let $\G\in\R^{\ssize \times \odim}$ be any matrix.
Let also $\Gr\in\R^{\ssize \times \rdim}$ be the 
sketch of $\G$ given by Top Outputs.
Then 
\begin{align*}
    \Error(\Sf_{\G},\Sf_{\Gr}) 
    \leq 
    \sum_{j = \rdim+1}^{\odim} \|\g_{i_j}\|^2.
\end{align*}
\end{proposition}
\begin{proof}
\Cref{th:mainbound} implies that
\begin{align*}
    \sup_{R}\bigl|\Sf_{\G}(R) - \Sf_{\Gr}(R)\bigr| 
    &\leq
    \bigl\|\G\G^{\T} - \Gr\Gr^{\T}\bigr\|.
\end{align*}
We rewrite $\G \G^{\T}$ and $\Gr \Gr^{\T}$ as an
outer product of their columns, 
\[
    \G \G^{\T} = \sum_{i=1}^{\odim} \g_i \g_i^{\T}
    \quad\text{and}\quad
    \Gr \Gr^{\T} = \sum_{j=1}^{\rdim} \g_{i_j} \g_{i_j}^{\T}.
\]
see Section~3.2 for notation. 
By construction, we have 
\begin{align*}
    \bigl\|\G\G^{\T} - \Gr \Gr^{\T}\bigr\| 
    = 
    \Biggr\|\sum_{j = \rdim+1}^{\odim}
    \g_{i_j} \g_{i_j}^{\T}\Biggr\|
    \leq
    \sum_{j = \rdim+1}^{\odim} 
    \bigl\|\g_{i_j}\bigr\|^2,
\end{align*}
and the proof is complete.
\end{proof}
This proposition shows that the approximation error 
is small when we cut out columns of $\G$ with a
small norm. 
Top Outputs method is less preferable 
than Truncated SVD in terms of the approximation error.
Nevertheless, here the sketch $\Gr$ 
can be computed in time $O(\ssize\odim)$
which is linear in $\odim$ contrary to the Truncated SVD.

\subsection{Random Sampling}

In Random Sampling, we
sample columns of $\G$
according to probabilities
proportional to their Euclidian norm.
Before we proceed, let us denote the stable rank of $G$ by 
\[
    \sr{\G}= \frac{\|\G\|^2_{\Fr}}{\|\G\|^2},
\]
where $\|\cdot\|$ denotes the spectral norm.
The stable rank is a relaxation of the exact notion of rank. 
Indeed, one always has $\sr{\G}\leq\operatorname{rank}(\G)$.
But as opposed to the exact rank, it is stable
under small perturbations of the matrix.
Both exact and stable ranks are usually
referred to as the intrinsic dimensionality of a matrix
(in data-driven applications matrices tend to have small ranks).
\begin{proposition}
\label{th:impsamplebound}
Let $\G\in\R^{\ssize \times \odim}$ be any matrix.
Let also $\Gr\in\R^{\ssize \times \rdim}$ be a sketch 
obtained by Random Sampling. 
Then for any $\delta \in (0,1)$, 
with probability at least $1-\delta$,
\begin{align*}
    \Error(\Sf_{\G},\Sf_{\Gr}) 
    \leq 
    C_{\G,\delta} \, \frac{\|\G\|^2}{\sqrt{\rdim}},
\end{align*}
where $C_{\G,\delta}$ is a constant 
depending on $\G$ and $\delta$ and is given by
\[
    C_{\G,\delta} = 2 \sqrt{\sr{G} \log\left(\frac{4\sr{G}}{\delta}\right)}.
\]
\end{proposition}
\begin{proof}
\Cref{th:mainbound} yields
\begin{align*}
    \sup_{R}\bigl|\Sf_{\G}(R) - \Sf_{\Gr}(R)\bigr| 
    &\leq
    \bigl\|\G\G^{\T} - \Gr\Gr^{\T}\bigr\|.
\end{align*}
Using Theorem~4.2 from \citep{holodnak-ipsen-2015}, 
we obtain that for any $\eps,\delta\in(0,1)$
with probability at least $1-\delta$,
\[
    \| \G\G^{\T} - \Gr\Gr^{\T} \| 
    \leq \eps \|\G\|^2
\]
provided that $\rdim \ge 3 \sr{G} \ln(\frac{4r}{\delta})/\eps^2$.
Solving the latter inequality with respect to $\eps$,
we establish the assertion.
\end{proof}

This proposition states that 
the approximation error is, 
with high probability, of order $\|\G\|^2/\sqrt{\rdim}$
when the stable rank of $\G$ is small.
There is no definite answer whether this bound 
is better than the bounds obtained for other methods. 
The answer depends on the spectrum of $\G$.
Moreover, this bound is of probabilistic nature.
Nevertheless, Random Sampling has the computational complexity $O(\ssize\odim)$ and is as fast as Top Outputs. 

\subsection{Random Projections}

Random Projections samples $\rdim$ random linear combinations 
of columns of G to construct $\Gr$.
\begin{proposition}
\label{th:jlbound}
Let $\G\in\R^{\ssize \times \odim}$ be any matrix.
Let also $\Pi\in\R^{\odim \times \rdim}$ be 
a random matrix filled with independently sampled $\NN(0,\rdim^{-1})$
entries. Set $\Gr = \G\Pi$.
Then for any $\delta \in (0,1)$, 
with probability at least $1-\delta$,
\begin{align*}
    \Error(\Sf_{\G},\Sf_{\Gr}) 
    \leq 
    C_{\G,\delta} \, \frac{\|\G\|^2}{\sqrt{\rdim}},
\end{align*}
where $C_{\G,\delta}$ is a constant depending on $\G$ and $\delta$  
and is given by
\[
    C_{\G,\delta} = c \sqrt{\sr{G} + \ln\left(\frac{1}{\delta}\right)}.
\]
for some absolute constant $c>0$.
\end{proposition}

\begin{proof}
Using \Cref{th:mainbound}, we get
\begin{align*}
    \sup_{R}\bigl|\Sf_{\G}(R) - \Sf_{\Gr}(R)\bigr| 
    &\leq
    \bigl\|\G\G^{\T} - \Gr\Gr^{\T}\bigr\|.
\end{align*}
Now Theorem~1 from \citep{kyrillidis-2014} 
implies that for any $\eps,\delta\in(0,1)$
with probability at least $1-\delta$,
\[
    \| \G\G^{\T} - \Gr\Gr^{\T} \| 
    \leq \eps \|\G\|^2
\]
provided that $\rdim \ge c \left(\sr{G} +\ln\ln(\frac{1}{\eps}) +\ln(\frac{1}{\delta})\right)/\eps^2$. If we set 
\[
    \eps = c' \sqrt{(\sr{G} +\ln(\tfrac{1}{\delta}))\big/\rdim}
\]
for another absolute constant $c'$, the assertion follows.
\end{proof}

Comparing to \Cref{th:impsamplebound}, this bound
is slightly better in terms of $C_{\G,\delta}$. 
But the sketch $G_{\rdim}$ here can be computed only in time 
$O(\ssize\odim\rdim)$ since it requires multiplication of 
$\G$ and $\Pi$. To speed up it, one
can use Fast JL transform \citep{ailon-chazelle-2009} or Sparse JL transform \citep*{dasgupta-2010}, \citep{kane-nelson-2014}.

\section{Experiment Details}
\label{sec:experiment_details}

We remind the reader that our Python-based GPU 
implementation of GBDT called
Py-Boost is available on 
\href{https://github.com/sb-ai-lab/Py-Boost}{GitHub}\footnote{\url{https://github.com/sb-ai-lab/Py-Boost}}.
The code to reproduce the experiments is 
also available on 
\href{https://github.com/sb-ai-lab/sketchboost-paper}{GitHub}\footnote{\url{https://github.com/sb-ai-lab/SketchBoost-paper}}. 

\subsection{About Py-Boost}
\label{sec:about_pyboost}

As was mentioned in the original paper, 
Py-Boost is written in Python 
and follows the classic 
scheme described in \citep{xgboost-2016}.
Meanwhile, it is a simplified version of gradient boosting, and hence it has a few limitations.
Some of these limitations have been made to 
speed up computations, some --- 
to remove unnecessary for our purposes 
features presented in modern 
gradient boosting toolkits
(for example, categorical data handling).
The complete list of these limitations is the following.
Py-Boost supports:
(a) computations only on GPU, 
(b) only the depth-wise tree growth policy,
(c) only numeric features (with possibly NaN values),
and
(d) only histogram algorithm for split search 
    (maximum number of bins for each feature is limited to $256$).
\par
Py-Boost uses GPU Python libraries such as CuPy, Numba, and CuML to speed up computations. XGBoost and CatBoost frameworks are also evaluated in the GPU mode where possible
(CatBoost is evaluated on CPU on multilabel classification and multioutput regression tasks since it does not support them on GPU).

\subsection{Experiment design}
\label{sec:experiment_design}

In our numerical experiments, we compare SketchBoost Full, 
SketchBoost with sketching strategies (Top Outputs, Random Sampling, Random Projections), 
XGBoost (v1.6.0) which uses the one-vs-all strategy,
and
CatBoost (v1.0.5) which uses the single-tree strategy,
and a popular deep learning model
for tabular data TabNet (v3.1.1).
In this section, we discuss experiment design for GBDTs;
details for TabNet are given in \Cref{sec:experiment_design_tabnet} below. 
The experiments are conducted on $9$ 
real-world publicly available 
datasets from Kaggle, OpenML, and Mulan website
for multiclass/multilabel classification and multitask regression.
Datasets details are given in \Cref{tb:datasets}. 
\par
\begin{table}[htbp]
  \setlength\tabcolsep{3pt}
  \setlength\extrarowheight{2pt}
  \centering
  \caption{Dataset statistics.\label{tb:datasets}}
  \scalebox{0.85}{
  \begin{tabular}{@{\extracolsep{4pt}}lcrrrcc@{}}
    \toprule
    \textbf{Dataset}  & \textbf{Task} & \textbf{Rows} & \textbf{Features} & \textbf{Classes/Labels/Targets} & \textbf{Source}& \textbf{Download} \\
    \midrule
    Otto & multiclass & 61\,878 & 93 & 9 & Kaggle & \href{https://www.kaggle.com/c/otto-group-product-classification-challenge}{\underline{Link}}\\
    SF-Crime & multiclass & 878\,049 & 10 & 39 & Kaggle & 
    \href{https://www.kaggle.com/c/sf-crime}{\underline{Link}}\\
    Helena & multiclass & 65\,196 & 27 & 100 & OpenML &
    \href{https://www.openml.org/d/41167}{\underline{Link}}\\
    Dionis & multiclass & 416\,188 & 60 & 355 & OpenML & 
    \href{https://www.openml.org/d/41169}{\underline{Link}}\\
    \midrule
    Mediamill & multilabel & 43\,907 & 120 & 101 & Mulan & \href{http://mulan.sourceforge.net/datasets-mlc.html}{\underline{Link}}\\
    MoA & multilabel & 23\,814 & 876 & 206 & Kaggle & 
    \href{https://www.kaggle.com/c/lish-moa}{\underline{Link}}\\
    Delicious & multilabel & 16\,105 & 500 & 983 & Mulan & \href{http://mulan.sourceforge.net/datasets-mlc.html}{\underline{Link}}\\
    \midrule
    RF1 & multitask & 9125 & 64  & 8 & Mulan & \href{http://mulan.sourceforge.net/datasets-mtr.html}{\underline{Link}}\\
    SCM20D & multitask & 8966 & 61 & 16 & Mulan & \href{http://mulan.sourceforge.net/datasets-mtr.html}{\underline{Link}}\\
    \bottomrule
  \end{tabular}}
\end{table}
\par
We remind the reader that if there is no 
official train/test split,
we split the data into train and test sets with ratio 80\%-20\%.
Datasets taken from Kaggle have the official 
train/test split, but since the platform hides 
the test set, we split the train test into 
the new train and test sets.
Some of the datasets required data preprocessing 
since they contained categorical 
and datetime features 
(they cannot be handled by all of the GBDT implementations on the fly). 
The code for data preprocessing step is also 
available on GitHub.
\par
Experiments are performed on the server under OS Ubuntu 18.04 with 4 NVidia Tesla V100 32 GB GPUs, 48 cores CPU Intel(R) Xeon(R) Platinum 8168 CPU @ 2.70GHz and 386 GB RAM. 
We run all the tasks on 8 CPU threads and 
single GPU if needed.

\par
The experiments are divided into two the following parts:
\begin{itemize}[leftmargin=*]
    \item \textbf{Parameter tuning.} Here we optimize hyperparameters
    for XGBoost and CatBoost. For SketchBoost we use the same hyperparameters as for CatBoost (to speed up the parameter tuning process; we do not expect that hyperparameters will vary much since we use the same tree building strategy).
    At this stage, parameters are estimated by 5-fold cross-validation using only the train set.
    \item \textbf{Model evaluation.} 
    After the best parameters are found, 
    we refit all the models using a longer 
    training time.
    The models are trained again by 5-fold cross-validation,
    but their quality is estimated on the test set.
    The final score and time metrics are computed as 
    the average of 5 values given by 5 models 
    from the cross-validation loop.
\end{itemize}
During the parameter tuning, we use a slightly different setup (higher learning rate and less maximum number of rounds) to evaluate more trials and find a better hyperparameter set. 
\Cref{tb:exp_diff} provides the associate details.
\par
\begin{table}[htbp]
  \setlength\tabcolsep{3pt}
  \setlength\extrarowheight{2pt}
  \centering
  \caption{ Setup for the parameter tuning and model evaluation stages.\label{tb:exp_diff}}
  \scalebox{0.84}{
  \begin{tabular}{@{\extracolsep{4pt}}lcrrrcc@{}}
    \toprule
    \textbf{Stage}  & \textbf{Learning Rate} & \textbf{Max Number of Rounds} & \textbf{Early Stopping Rounds} & \textbf{Quality Estimation}  \\
    \midrule
    Parameter tuning & 0.05 & 5000  & 200 & cross-validation \\
    Models evaluation & 0.015 & 20000 & 500 & test set \\
    \bottomrule
  \end{tabular}}
\end{table}
\par
Since all the models are trained using cross-validation, the optimal number of boosting rounds is determined adaptively by early stopping on the validation fold. In this setup, the test set is used only for quality evaluation. 
As the primary quality measure, 
we use the cross-entropy for classification 
and RMSE for regression.
But for the sake of completeness, we also report
the accuracy for classification 
and R-squared for regression, see \Cref{sec:experiment_results}.

\subsection{Hyperparameter tuning}
\label{sec:experiment_hype}

It is quite challenging to perform a fair test among all the frameworks. Evaluation results depend not only on the sketching method or the strategy used to handle a multioutput problem (one-vs-all or single-tree), but also on hyperparameter setting and implementation details. To make our comparison as fair as possible, we perform a hyperparameter optimization 
for XGBoost and CatBoost using the 
Optuna\footnote{\url{https://github.com/optuna/optuna}} framework that performs a sequential model-based optimization by the Tree-structured Parzen Estimator (TPE) method; see  \citep*{bergstra2011algorithms}. 
The list of optimized hyperparameters is given in \Cref{tb:hyperlist}.
\par
\begin{table}[htbp]
  \setlength\tabcolsep{3pt}
  \setlength\extrarowheight{2pt}
  \centering
  \caption{ Optimized hyperparameters.\label{tb:hyperlist}}
  \scalebox{0.85}{
  \begin{tabular}{@{\extracolsep{4pt}}lcrrrcc@{}}
    \toprule
    \textbf{Parameter}  & \textbf{Framework} & \textbf{Type} & \textbf{Default value} & \textbf{Search space} & \textbf{Log}  \\
    \midrule
    Maximal Tree Depth & All & Int  & 6 & [3:12] & False \\
    Min Child Weight & XGBoost & Float & 1e-5 & [1e-5:10] & True \\
    Min Data In Leaf & CatBoost & Int & 1 & [1: 100] & True \\
    L2 leaf regularization & All & Float & 1 & [0.1: 50] & True \\
    Rows Sampling Rate & All & Float & 1.0 & [0.5: 1] & False\\
    \bottomrule
  \end{tabular}}
\end{table}
\par
\textbf{Note}: 
{Minimal data in leaf} and {Minimal child weight} both control the leaf size but in different frameworks. It is the reason why they have different scales in the table.
\par
The optimal number of rounds is determined by early stopping for the fixed learning rate; see \Cref{sec:experiment_design} for the details. There are some other parameters of GBDT that are not optimized since not all the frameworks allow to change them. For example, the columns sampling rate cannot be changed in CatBoost in the GPU mode, and only the depth-wise grow policy is supported in SketchBoost.
\par
The tuning process is organized in the following way:
\begin{itemize}[leftmargin=*]
    \item \textbf{Safe run.} First, we perform a single run with a set of parameters that are listed in \Cref{tb:hyperlist} as default. These hyperparameters are close to the framework's default settings and can be considered as a safe option in the case when a good set of parameters is not found in the time limit.
    \item \textbf{TPE search.} Then we perform 30 iterations of parameters search with the Optuna framework. As it was mentioned, the training time may be quite long for the multioutput tasks, so we limit the search time to 24 hours in order to finish it in a reasonable amount of time.
    \item \textbf{Selection.} Finally, we determine the best parameters as the ones that achieve the best performance between all trials (both safe and TPE).
\end{itemize}
The best hyperparameters and the number of successful (finished in time) trials are given in \Cref{tb:hyperparameters}.
\par
\begin{table}[htbp]
  \setlength\tabcolsep{3pt}
  \setlength\extrarowheight{2pt}
  \captionsetup{justification=centering}
  \centering
  \caption{
  Results for hyperparameter optimization.
  \label{tb:hyperparameters}}
  \scalebox{0.85}{
  \begin{tabular}{@{\extracolsep{4pt}}llcccccc@{}}
    \toprule
    \textbf{Dataset} & \textbf{Framework}  & 
    \textbf{Min} & \textbf{Min} & \textbf{Rows sampling} & \textbf{Max} & 
    \textbf{L2 leaf} & \textbf{Completed}\\
    & & \textbf{data in leaf} & \textbf{child weight} & \textbf{rate}&\textbf{depth} &\textbf{regularization} & \textbf{trials} \\
    \midrule
    Otto & CatBoost & 47 & -- & 0.89 &  10 & 3.83 & 31 \\
     & XGBoost & -- & 0.00001 & 0.58 & 12 & 23.57 & 31 \\
    \midrule
    SF-Crime & CatBoost & 2 & -- & 0.94 & 11 & 2.65 & 31  \\
     & XGBoost & -- & 1.25682 & 0.92 & 12 & 8.37 & 24 \\
    \midrule
    Helena & CatBoost & 2 & -- & 0.55 & 6 & 12.77 & 31 \\
     & XGBoost & -- & 0.33734 & 0.50 & 6 & 23.33 & 31 \\
    \midrule
    Dionis & CatBoost & 1 & -- & 1.0 & 6 & 1.0 & 6\\
     & XGBoost & -- & 0.00001 & 1.0 & 6 & 1.0 & 3\\
    \midrule
    Mediamill & CatBoost & 10 & -- & 0.76 & 8 & 11.16 & 8\\
     & XGBoost & -- & 1.90469 & 0.68 & 12 & 39.30 & 31\\
    \midrule
    MoA & CatBoost & 30 & -- & 0.88 & 4 & 1.03 & 10 \\
     & XGBoost & -- & 0.00342 & 0.93 & 3 & 0.37 & 31\\
    \midrule
    Delicious & CatBoost & 76 & -- & 0.62 & 12 & 13.05 & 5\\
     & XGBoost & -- & 0.06235 & 0.72 & 11 & 33.22 & 10\\
    \midrule
    RF1 & CatBoost & 1 & -- & 0.85 & 10 & 9.96 & 31\\
     & XGBoost & -- & 0.00431 & 0.68 & 5 & 12.22 & 31\\
    \midrule
    SCM20D & CatBoost & 5 & -- & 0.99 & 12 & 6.57 & 31\\
     & XGBoost & -- & 9.93770 & 0.80 & 6 & 1.44 & 31\\
    \bottomrule
  \end{tabular}}
\end{table}

\subsection{Details on TabNet}
\label{sec:experiment_design_tabnet}
Here we provide additional details on TabNet \citep{arik2021tabnet}. We consider PyTorch (v1.7.0)-based implementation pytorch-tabnet library (v3.1.1). The evaluation is performed on the same datasets, under the same experiment setup including train/validation/test splits, and using the same hardware as described in \Cref{sec:experiment_design}.
TabNet hyperparameters are taken from the original paper \citep{arik2021tabnet} except the learning rate, batch size, and the number of epochs for train and early stopping. These hyperparameters are tuned in the following way:

\begin{itemize}[leftmargin=*]
    \item The optimal learning rate is tuned using the Optuna framework from the range ($1e-5$; $1e-1$) in the log scale
    in the same way at it is described in \Cref{sec:experiment_hype} (one single safe run with learning rate $2e-2$ and then TPE search for $10$ trials with time limit of $24$ hours). The best trial is selected based on the cross-validation score and then it is evaluated on the test set. 
    \item The optimal number of epochs is selected via early stopping with $16$ epochs with no improve. Maximal number of epochs is limited to $500$.
    \item The batch size is selected depending on the dataset size: $256$ for data with less than $50$k rows, $512$ for $50$-$100$k rows, and $1024$ for more then $100$k rows. 
\end{itemize}

The proposed values are quite typical for tabular data according to \citep*{gorishniy2021revisiting}, \citep{fiedler2021simple}, and \citep{arik2021tabnet}. The unsupervised pretrain proposed in the paper is also used on each cross-validation iteration. The selected hyperparameters are listed in the \Cref{tb:tabnet_params} below.

\begin{table}[htbp]
  \setlength\tabcolsep{3pt}
  \setlength\extrarowheight{2pt}
  \captionsetup{justification=centering}
  \centering
  \caption{ 
  Results for hyperparameter optimization for TabNet.
  \label{tb:tabnet_params}}
  \scalebox{0.85}{
  \begin{tabular}{@{\extracolsep{4pt}}llcccccc@{}}
    \toprule
    \textbf{Dataset} &  
    \textbf{Learning} & \textbf{Batch} & \textbf{Completed}\\
    & \textbf{rate} & \textbf{size}& \textbf{trials} \\
    \midrule
    Otto & 0.0394 & 512 & 11 \\
    \midrule
    SF-Crime  & 0.0200 & 1024  & 3  \\
    \midrule
    Helena & 0.0080 & 512 & 9 \\
    \midrule
    Dionis & 0.0110  & 1024 & 7 \\
    \midrule
    Mediamill  & 0.0315 & 256 & 11 \\
    \midrule
    MoA  & 0.0200 & 256 &  10 \\
    \midrule
    Delicious & 0.0231 & 256 & 11 \\
    \midrule
    RF1 & 0.0200 & 256 & 11 \\
    \midrule
    SCM20D  & 0.0200 & 256  & 11 \\
    \bottomrule
  \end{tabular}}
\end{table}

\clearpage
\subsection{Additional experimental results}
\label{sec:experiment_results}

\begin{figure}[ht!]
\captionsetup{justification=centering}
\centering
\includegraphics[width=0.3\linewidth]{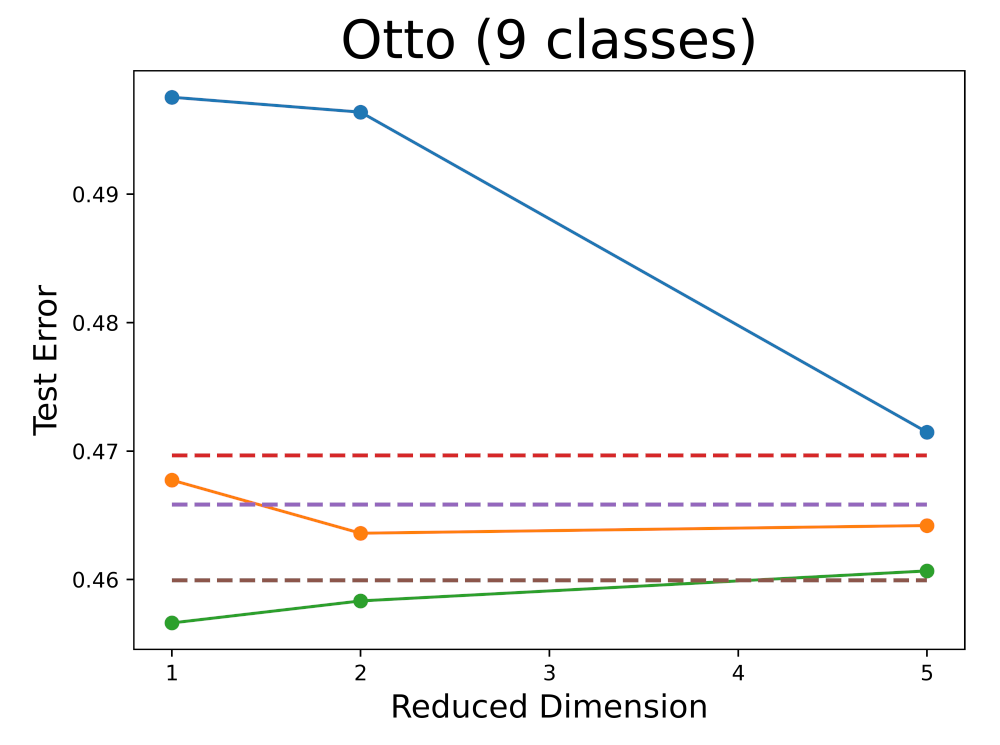}
\includegraphics[width=0.3\linewidth]{pics/test_score_mini/sf-crime.png}
\includegraphics[width=0.3\linewidth]{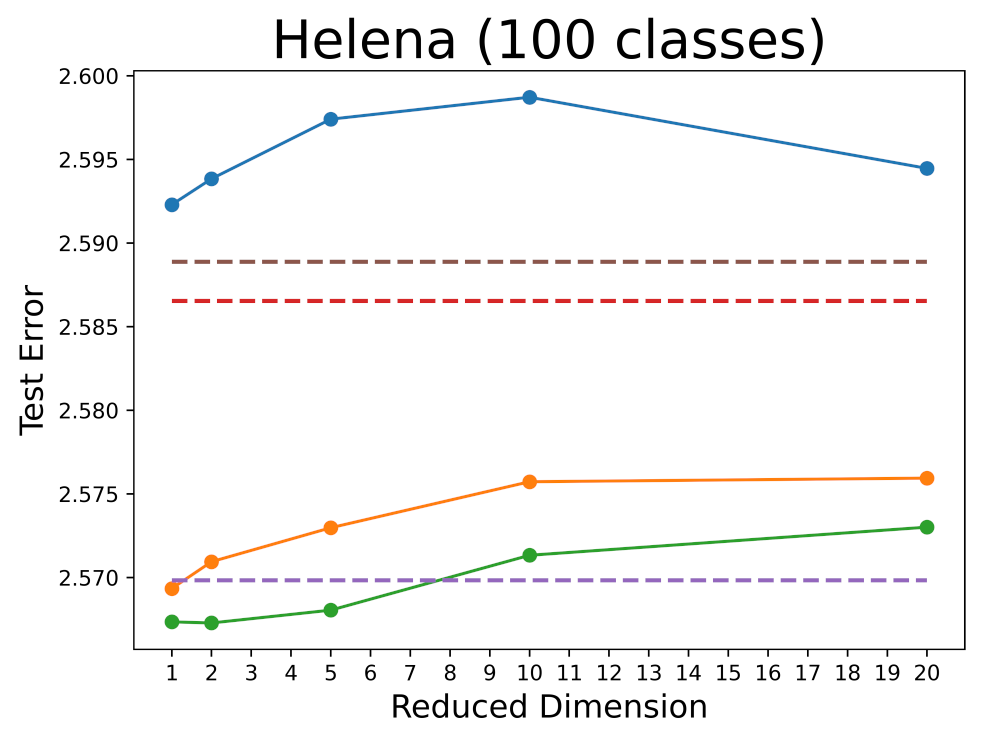}
\\
\includegraphics[width=0.3\linewidth]{pics/test_score_mini/dionis.png}
\includegraphics[width=0.3\linewidth]{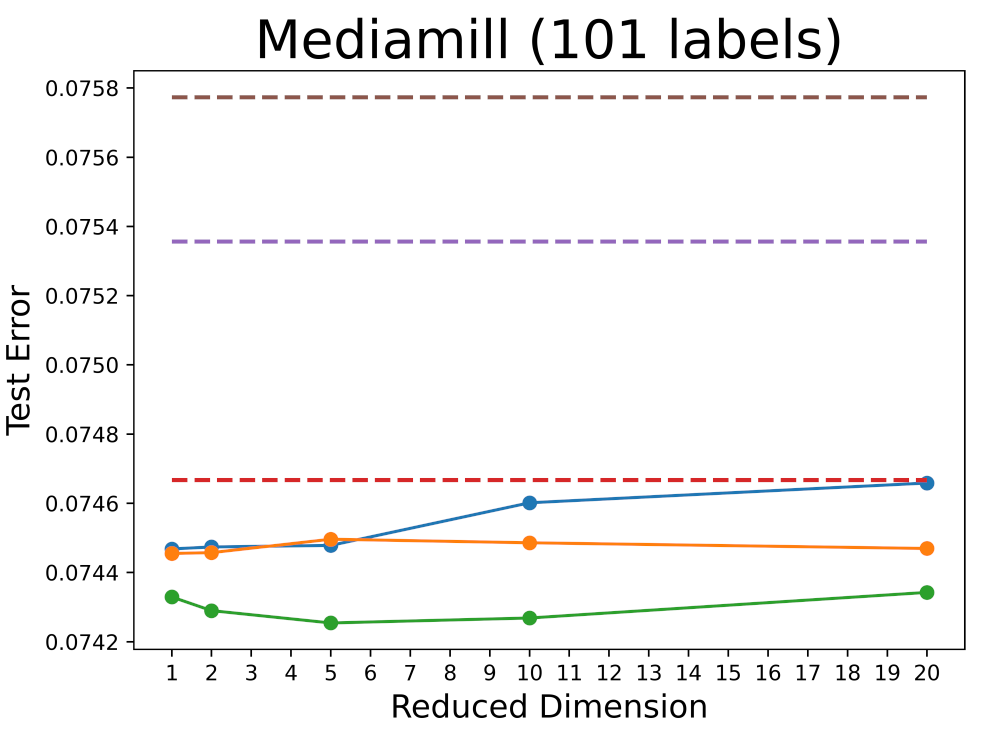}
\includegraphics[width=0.3\linewidth]{pics/test_score_mini/moa.png}
\\
\includegraphics[width=0.3\linewidth]{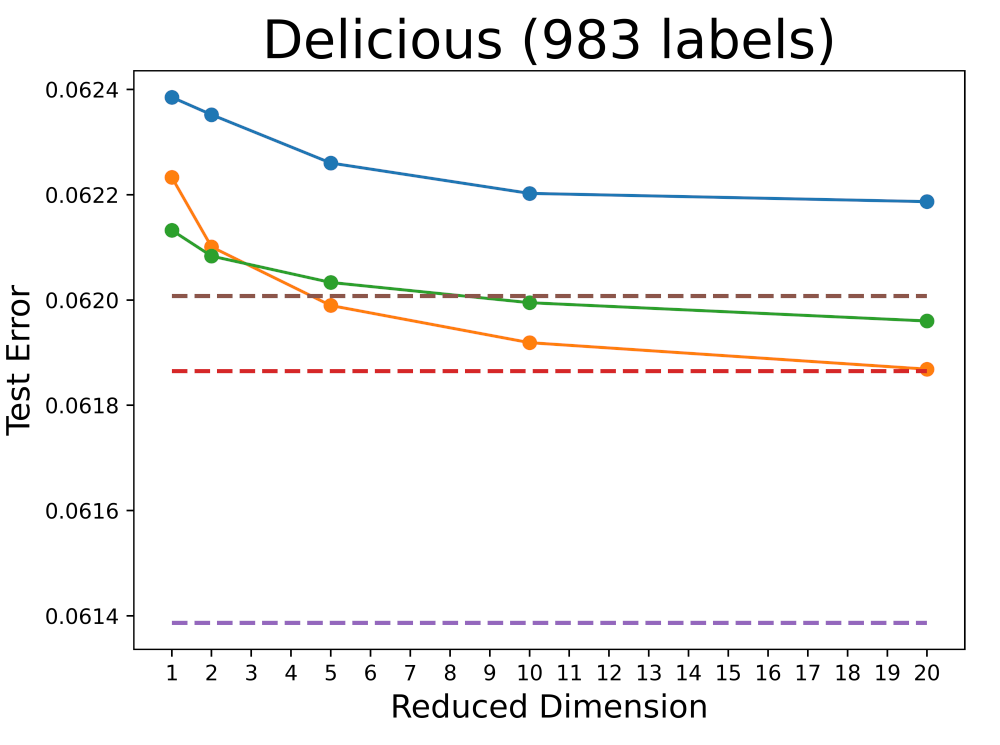}
\includegraphics[width=0.3\linewidth]{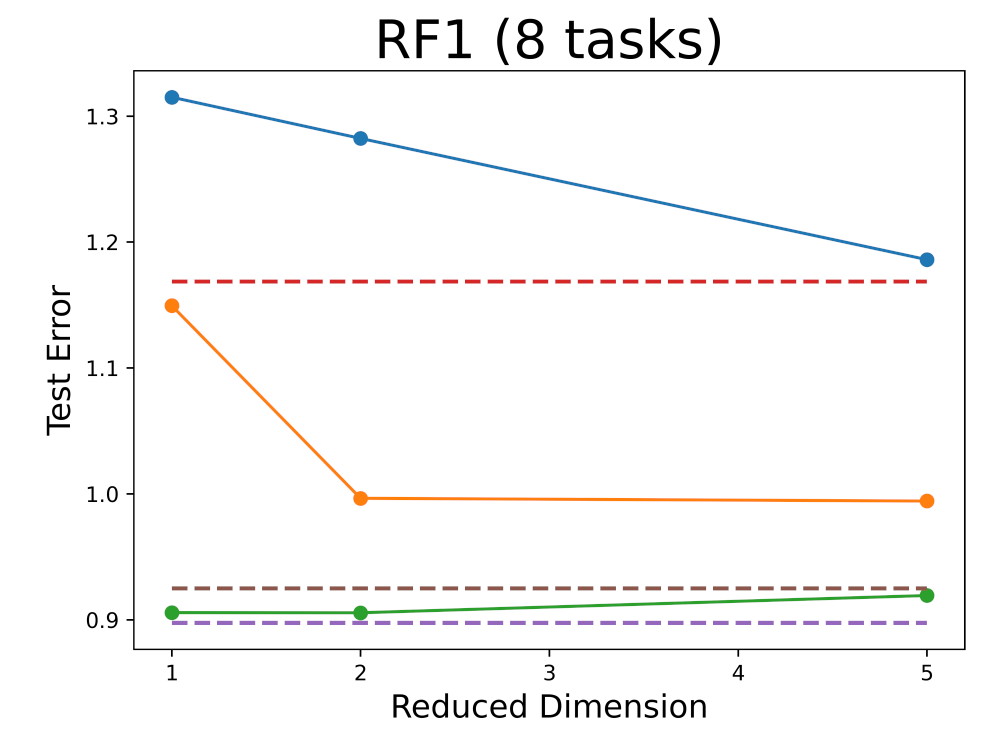}
\includegraphics[width=0.3\linewidth]{pics/test_score_mini/scm20d.png}
\\
\includegraphics[width=0.95\linewidth]{pics/legend.png}
\caption{ Dependence of test errors (cross-entropy for classification and RMSE for regression) \\ on sketch dimension $\rdim$ for all datasets.\label{pic:test_score_appendix}}
\end{figure}

\begin{table}[htbp]
  \setlength\tabcolsep{3pt}
  \setlength\extrarowheight{2pt}
  \captionsetup{justification=centering}
  \centering
  \caption{ \small
  Test errors (cross-entropy for classification and RMSE for regression)
  $\pm$ their standard deviation.
  \label{tb:detailed_test_score_bce}}
  \scalebox{0.55}{
  \begin{tabular}{@{\extracolsep{4pt}}llccccccccc@{}}
    \toprule
    & &
    \multicolumn{9}{c}{\textbf{Dataset}} 
    \vspace{0.4em}\\
    \cline{3-11} \vspace{-0.2em}
    \textbf{Algorithm} & \textbf{}  & 
    \textbf{Otto} & \textbf{SF-Crime} & \textbf{Helena} & 
    \textbf{Dionis} & \textbf{Mediamill} & \textbf{MoA} & \textbf{Delicious} & \textbf{RF1} & \textbf{SCM20D}\\
    \midrule
    XGBoost & & 0.4599 & 2.2208 & 2.5889 & 0.3502 & 0.0758 & 0.0166 & 0.0620 & 0.9250 & 89.1045\\
     & & \qquad$\pm$0.0027 & \qquad$\pm$0.0008 & \qquad$\pm$0.0031 & \qquad$\pm$0.0019 & \qquad$\pm$1.1e-04 & \qquad$\pm$2.1e-05 & \qquad$\pm$3.3e-05 & \qquad$\pm$0.0307 & \qquad$\pm$0.4950\\
    CatBoost  & & 0.4658 & 2.2036 & 2.5698 & 0.3085 & 0.0754 & 0.0161 & 0.0614 & 0.8975 & 90.9814\\
     & & \qquad$\pm$0.0032 & \qquad$\pm$0.0005 & \qquad$\pm$0.0025 & \qquad$\pm$0.0010 & \qquad$\pm$1.1e-04 & \qquad$\pm$2.6e-05 & \qquad$\pm$5.2e-05 & \qquad$\pm$0.0383 & \qquad$\pm$0.3652\\
    TabNet  & & 0.5363 & 2.4819 & 2.7197 & 0.4753 & 0.0859 & 0.0193 & 0.0664 & 3.7948 & 87.3655\\
     & & \qquad$\pm$0.0063 & \qquad$\pm$0.0199 & \qquad$\pm$0.0235 & \qquad$\pm$0.0126 & \qquad$\pm$3.3e-03 & \qquad$\pm$3.0e-04 & \qquad$\pm$8.0e-04 & \qquad$\pm$1.5935 & \qquad$\pm$1.3316\\ 
    SketchBoost Full & & 0.4697 & 2.2067 & 2.5865 & 0.3114 & 0.0747 & 0.0160 & 0.0619 & 1.1687 & 91.0142\\
     & & \qquad$\pm$0.0029 & \qquad$\pm$0.0003 & \qquad$\pm$0.0025 & \qquad$\pm$0.0008 & \qquad$\pm$1.3e-04 & \qquad$\pm$9.0e-06 & \qquad$\pm$5.5e-05 & \qquad$\pm$0.0835 & \qquad$\pm$0.3396\\
    \midrule
    Top Outputs & $k=1$ & 0.4975 & 2.2282 & 2.5923 & 0.3339 & 0.0745 & 0.0169 & 0.0624 & 1.3151 & 90.7613\\
    & & \qquad$\pm$0.0030 & \qquad$\pm$0.0004 & \qquad$\pm$0.0024 & \qquad$\pm$0.0017 & \qquad$\pm$1.3e-04 & \qquad$\pm$4.1e-05 & \qquad$\pm$6.3e-05 & \qquad$\pm$0.0721 & \qquad$\pm$0.3988\\
    & $k=2$ & 0.4964 & 2.2284 & 2.5938 & 0.3217 & 0.0745 & 0.0168 & 0.0624 & 1.2823 & 89.5284\\
    &  & \qquad$\pm$0.0041 & \qquad$\pm$0.0001 & \qquad$\pm$0.0025 & \qquad$\pm$0.0019 & \qquad$\pm$1.4e-04 & \qquad$\pm$2.0e-05 & \qquad$\pm$5.1e-05 & \qquad$\pm$0.1363 & \qquad$\pm$0.8352\\
    & $k=5$ & 0.4715 & 2.2183 & 2.5974 & 0.3155 & 0.0745 & 0.0168 & 0.0623 & 1.1860 & 88.7442\\
    &  & \qquad$\pm$0.0035 & \qquad$\pm$0.0005 & \qquad$\pm$0.0018 & \qquad$\pm$0.0013 & \qquad$\pm$1.1e-04 & \qquad$\pm$1.6e-05 & \qquad$\pm$6.3e-05 & \qquad$\pm$0.1365 & \qquad$\pm$0.6345\\
    & $k=10$ & -- & 2.2116 & 2.5987 & 0.3151 & 0.0746 & 0.01660 & 0.0622 & -- & 89.8727\\
    & & -- & \qquad$\pm$0.0025 & \qquad$\pm$0.0019 & \qquad$\pm$0.0014 & \qquad$\pm$1.0e-04 & \qquad$\pm$2.5e-05 & \qquad$\pm$5.5e-05 & -- & \qquad$\pm$0.3126\\
    & $k=20$ & -- & 2.2070 & 2.5945 & 0.3146 & 0.0747 & 0.0163 & 0.0622 & -- & --\\
    & & -- & \qquad$\pm$0.0005 & \qquad$\pm$0.0020 & \qquad$\pm$0.0010 & \qquad$\pm$1.1e-04 & \qquad$\pm$2.2e-05 & \qquad$\pm$6.2e-05 & -- & --\\
    \midrule
    Random Sampling & $k=1$ & 0.4677 & 2.2140 & 2.5693 & 0.3175 & 0.0745 & 0.0163 & 0.0622 & 1.1495 & 87.9358\\
    & & \qquad$\pm$0.0019 & \qquad$\pm$0.0003 & \qquad$\pm$0.0022 & \qquad$\pm$0.0009 & \qquad$\pm$1.3e-04 & \qquad$\pm$2.0e-05 & \qquad$\pm$6.4e-05 & \qquad$\pm$0.0674 & \qquad$\pm$0.4111\\
    & $k=2$ & 0.4636 & 2.2083 & 2.5710 & 0.3089 & 0.0745 & 0.0162 & 0.0621 & 0.9965 & 86.7842\\
    & & \qquad$\pm$0.0025 & \qquad$\pm$0.0003 & \qquad$\pm$0.0032 & \qquad$\pm$0.0012 & \qquad$\pm$9.1e-05 & \qquad$\pm$1.5e-05 & \qquad$\pm$5.5e-05 & \qquad$\pm$0.1011 & \qquad$\pm$0.5546\\
    & $k=5$ & 0.4642 & 2.2052 & 2.5730 & 0.3055 & 0.0745 & 0.0161 & 0.0620 & 0.9944 & 86.2964\\
    & & \qquad$\pm$0.0020 & \qquad$\pm$0.0005 & \qquad$\pm$0.0024 & \qquad$\pm$0.0012 & \qquad$\pm$9.8e-05 & \qquad$\pm$1.5e-05 & \qquad$\pm$5.6e-05& \qquad$\pm$0.1014 & \qquad$\pm$0.4398\\
    & $k=10$ & -- & 2.2041 & 2.5757 & 0.3051 & 0.0745 & 0.0161 & 0.0619 & -- & 86.6865\\
    & & -- & \qquad$\pm$0.0005 & \qquad$\pm$0.0018 & \qquad$\pm$0.0009 & \qquad$\pm$1.0e-04 & \qquad$\pm$1.5e-05 & \qquad$\pm$5.2e-05 & -- & \qquad$\pm$0.2829\\
    & $k=20$ & -- & 2.2037 & 2.5759 & 0.3040 & 0.0745 & 0.0160 & 0.0619 & -- & --\\
    & & -- & \qquad$\pm$0.0004 & \qquad$\pm$0.0023 & \qquad$\pm$0.0013 & \qquad$\pm$1.1e-04 & \qquad$\pm$1.0e-05 & \qquad$\pm$5.9e-05 & -- & --\\
    \midrule
    Random Projection & $k=1$ & 0.4566 & 2.2096 & 2.5674 & 0.2848 & 0.0743 & 0.0160 & 0.0621 & 0.9058 & 86.1442\\
    & & \qquad$\pm$0.0023 & \qquad$\pm$0.0005 & \qquad$\pm$0.0039 & \qquad$\pm$0.0012 & \qquad$\pm$1.1e-04 & \qquad$\pm$1.0e-05 & \qquad$\pm$5.9e-05 & \qquad$\pm$0.0442 & \qquad$\pm$0.4824\\
    & $k=2$ & 0.4583 & 2.2076 & 2.5673 & 0.2850 & 0.0743 & 0.0160 & 0.0621 & 0.9056 & 85.8061\\
    & & \qquad$\pm$0.0028 & \qquad$\pm$0.0006 & \qquad$\pm$0.0026 & \qquad$\pm$0.0011 & \qquad$\pm$1.2e-04 & \qquad$\pm$2e-05 & \qquad$\pm$5.9e-05 & \qquad$\pm$0.0581 & \qquad$\pm$0.5533\\
    & $k=5$ & 0.4607 & 2.2052 & 2.5681 & 0.2866 & 0.0743 & 0.0160 & 0.0620 & 0.9193 & 85.8565\\
    & & \qquad$\pm$0.0031 & \qquad$\pm$0.0006 & \qquad$\pm$0.0019 & \qquad$\pm$0.0013 & \qquad$\pm$1.2e-04 & \qquad$\pm$1.9e-05 & \qquad$\pm$5.6e-05 & \qquad$\pm$0.0781 & \qquad$\pm$0.3116\\
    & $k=10$ & -- & 2.2043 & 2.5713 & 0.2881 & 0.0743 & 0.0160 & 0.0620 & - & 86.3126\\
    & & -- & \qquad$\pm$0.0003 & \qquad$\pm$0.0024 & \qquad$\pm$0.0010 & \qquad$\pm$1.1e-04 & \qquad$\pm$1.3e-05 & \qquad$\pm$5.9e-05 & - & \qquad$\pm$0.3710\\
    & $k=20$ & -- & 2.2038 & 2.5730 & 0.2907 & 0.0743 & 0.0160 & 0.062 & -- & --\\
    & & -- & \qquad$\pm$0.0004 & \qquad$\pm$0.0030 & \qquad$\pm$0.0009 & \qquad$\pm$1.2e-04 & \qquad$\pm$6.0e-06 & \qquad$\pm$6.2e-05 & -- & --\\
    \bottomrule
  \end{tabular}}
\end{table}

\begin{table}[htbp]
  \setlength\tabcolsep{3pt}
  \setlength\extrarowheight{2pt}
  \captionsetup{justification=centering}
  \centering
  \caption{ \small
  Test errors (accuracy for classification and R-squared for regression)
  $\pm$ their standard deviation.
  \label{tb:detailed_test_score_acc}}
  \scalebox{0.55}{
  \begin{tabular}{@{\extracolsep{4pt}}llccccccccc@{}}
    \toprule
    & &
    \multicolumn{9}{c}{\textbf{Dataset}} 
    \vspace{0.4em}\\
    \cline{3-11} \vspace{-0.2em}
    \textbf{Algorithm} & \textbf{}  & 
    \textbf{Otto} & \textbf{SF-Crime} & \textbf{Helena} & 
    \textbf{Dionis} & \textbf{Mediamill} & \textbf{MoA} & \textbf{Delicious} & \textbf{RF1} & \textbf{SCM20D}\\
    \midrule
    XGBoost & & 0.8238 & 0.3326 & 0.3770 & 0.9193 & 0.9746 & 0.9971 & 0.9826 & 0.9997 & 0.9257\\
     & & \qquad$\pm$0.0010 & \qquad$\pm$0.0003 & \qquad$\pm$0.0012 & \qquad$\pm$0.0007 & \qquad$\pm$4.3e-05& \qquad$\pm$8.0e-06 & \qquad$\pm$6.0e-06 & \qquad$\pm$3.2e-05 & \qquad$\pm$0.0007\\
    CatBoost & & 0.8213 & 0.3352 & 0.3808 & 0.9234 & 0.9744 & 0.9971 & 0.9825 & 0.9997 & 0.9224\\
     & & \qquad$\pm$0.0012 & \qquad$\pm$0.0008 & \qquad$\pm$0.0017 & \qquad$\pm$0.0003 & \qquad$\pm$7.6e-05 & \qquad$\pm$5.0e-06 & \qquad$\pm$1.7e-05 & \qquad$\pm$4.1e-05 & \qquad$\pm$0.0006\\
    TabNet & & 0.7972 & 0.2550 & 0.3503 & 0.8936 & 0.9709 & 0.9967 & 0.9816 & 0.9932 & 0.9281 \\
     & & \qquad$\pm$0.0030 & \qquad$\pm$0.0037 & \qquad$\pm$0.0060 & \qquad$\pm$0.0032 & \qquad$\pm$1.0e-03 & \qquad$\pm$5.3e-05 & \qquad$\pm$9.5e-05 & \qquad$\pm$0.0037 & \qquad$\pm$0.0022\\ 
    SketchBoost Full & & 0.8223 & 0.3343 & 0.3783 & 0.9227 & 0.9747 & 0.9971 & 0.9824 & 0.9995 & 0.9224\\
     & & \qquad$\pm$0.0021 & \qquad$\pm$0.0007 & \qquad$\pm$0.0011 & \qquad$\pm$0.0004 & \qquad$\pm$6.8e-05 & \qquad$\pm$6.0e-06 & \qquad$\pm$1.4e-05 & \qquad$\pm$5.5e-05 & \qquad$\pm$0.0007\\
    \midrule
    Top Outputs & $k=1$ & 0.8172 & 0.3275 & 0.3773 & 0.9192 & 0.9747 & 0.9970 & 0.9823 & 0.9992 & 0.9228\\
    & & \qquad$\pm$0.0022 & \qquad$\pm$0.0006 & \qquad$\pm$0.0014 & \qquad$\pm$0.0006 & \qquad$\pm$4.9e-05 & \qquad$\pm$7.0e-06 & \qquad$\pm$2.0e-05 & \qquad$\pm$6.1e-05 & \qquad$\pm$0.0006\\
    & $k=2$ & 0.8171 & 0.3275 & 0.3772 & 0.9214 & 0.9747 & 0.9970 & 0.9823 & 0.9993 & 0.9249\\
    &  & \qquad$\pm$0.0016 & \qquad$\pm$0.0008 & \qquad$\pm$0.0021 & \qquad$\pm$0.0004 & \qquad$\pm$4.7e-05 & \qquad$\pm$4.0e-06 & \qquad$\pm$2.0e-05 & \qquad$\pm$1.2e-04 & \qquad$\pm$0.0013\\
    & $k=5$ & 0.8210 & 0.3315 & 0.3760 & 0.9229 & 0.9747 & 0.9970 & 0.9823 & 0.9994 & 0.9262\\
    & & \qquad$\pm$0.0016 & \qquad$\pm$0.0003 & \qquad$\pm$0.0019 & \qquad$\pm$0.0004 & \qquad$\pm$5.8e-05 & \qquad$\pm$1.2e-05 & \qquad$\pm$1.9e-05 & \qquad$\pm$9.2e-05 & \qquad$\pm$0.0010\\
    & $k=10$ & -- & 0.3333 & 0.3757 & 0.9229 & 0.9747 & 0.997 & 0.9824 & -- & 0.9243\\
    & & -- & \qquad$\pm$0.0003 & \qquad$\pm$0.0013 & \qquad$\pm$0.0003 & \qquad$\pm$2.8e-05 & \qquad$\pm$5.0e-06 & \qquad$\pm$1.5e-05 & -- & \qquad$\pm$0.0005\\
    & $k=20$ & -- & 0.3349 & 0.3766 & 0.9227 & 0.9747 & 0.9971 & 0.9824 & -- & --\\
    & & -- & \qquad$\pm$0.0006 & \qquad$\pm$0.0008 & \qquad$\pm$0.0007 & \qquad$\pm$6.5e-05 & \qquad$\pm$9.0e-06 & \qquad$\pm$1.6e-05 & -- & --\\
    \midrule
    Random Sampling & $k=1$ & 0.8228 & 0.3320 & 0.3821 & 0.9224 & 0.9746 & 0.9971 & 0.9824 & 0.9995 & 0.9276\\
    & & \qquad$\pm$0.0011 & \qquad$\pm$0.0002 & \qquad$\pm$0.0009 & \qquad$\pm$0.0005 & \qquad$\pm$5.6e-05 & \qquad$\pm$5e-06 & \qquad$\pm$2.0e-05 & \qquad$\pm$5.4e-05 & \qquad$\pm$0.0007\\
    & $k=2$ & 0.8236 & 0.3338 & 0.3827 & 0.9243 & 0.9746 & 0.9971 & 0.9824 & 0.9996 & 0.9293\\
    & & \qquad$\pm$0.0025 & \qquad$\pm$0.0003 & \qquad$\pm$0.0015 & \qquad$\pm$0.0003 & \qquad$\pm$4.8e-05 & \qquad$\pm$5.0e-06 & \qquad$\pm$8.0e-06 & \qquad$\pm$6.3e-05 & \qquad$\pm$0.0009\\
    & $k=5$ & 0.8231 & 0.3348 & 0.3795 & 0.9250 & 0.9746 & 0.9971 & 0.9824 & 0.9996 & 0.9301\\
    & & \qquad$\pm$0.0018 & \qquad$\pm$0.0002 & \qquad$\pm$0.0022 & \qquad$\pm$0.0002 & \qquad$\pm$4.0e-05 & \qquad$\pm$1.2e-05 & \qquad$\pm$1.3e-05 & \qquad$\pm$6.0e-05 & \qquad$\pm$0.0007\\
    & $k=10$ & -- & 0.3351 & 0.3793 & 0.9250 & 0.9746 & 0.9972 & 0.9824 & -- & 0.9295\\
    & & -- & \qquad$\pm$0.0004 & \qquad$\pm$0.0018 & \qquad$\pm$0.0002 & \qquad$\pm$4.1e-05 & \qquad$\pm$1.2e-05 & \qquad$\pm$8.0e-06 & -- & \qquad$\pm$0.0004\\
    & $k=20$ & -- & 0.3353 & 0.3795 & 0.9251 & 0.9746 & 0.9972 & 0.9824 & -- & --\\
    & & -- & \qquad$\pm$0.0005 & \qquad$\pm$0.0017 & \qquad$\pm$0.0005 & \qquad$\pm$5.6e-05 & \qquad$\pm$9.0e-06 & \qquad$\pm$1.7e-05 & -- & --\\
    \midrule
    Random Projection & $k=1$ & 0.8258 & 0.3338 & 0.3834 & 0.9285 & 0.9747 & 0.9971 & 0.9824 & 0.9997 & 0.9304\\
    & & \qquad$\pm$0.0010 & \qquad$\pm$0.0003 & \qquad$\pm$0.0033 & \qquad$\pm$0.0004 & \qquad$\pm$1.7e-05 & \qquad$\pm$6.0e-06 & \qquad$\pm$1.6e-05 & \qquad$\pm$3.7e-05 & \qquad$\pm$0.0008\\
    & $k=2$ & 0.8255 & 0.3344 & 0.3836 & 0.9287 & 0.9748 & 0.9971 & 0.9824 & 0.9997 & 0.9309\\
    & & \qquad$\pm$0.0023 & \qquad$\pm$0.0005 & \qquad$\pm$0.0019 & \qquad$\pm$0.0002 & \qquad$\pm$5.6e-05 & \qquad$\pm$3.0e-06 & \qquad$\pm$1.8e-05 & \qquad$\pm$4.4e-05 & \qquad$\pm$0.0008\\
    & $k=5$ & 0.8251 & 0.3351 & 0.3835 & 0.9281 & 0.9748 & 0.9971 & 0.9824 & 0.9997 & 0.9308\\
    & & \qquad$\pm$0.0023 & \qquad$\pm$0.0005 & \qquad$\pm$0.0018 & \qquad$\pm$0.0002 & \qquad$\pm$4.0e-05 & \qquad$\pm$1.1e-05 & \qquad$\pm$7.0e-06 & \qquad$\pm$5.4e-05 & \qquad$\pm$0.0005\\
    & $k=10$ & -- & 0.3355 & 0.3824 & 0.9279 & 0.9748 & 0.9971 & 0.9824 & - & 0.9301\\
    & & -- & \qquad$\pm$0.0003 & \qquad$\pm$0.0022 & \qquad$\pm$0.0002 & \qquad$\pm$4.0e-05 & \qquad$\pm$6.0e-06 & \qquad$\pm$9.0e-06 & - & \qquad$\pm$0.0006\\
    & $k=20$ & -- & 0.3357 & 0.3805 & 0.9275 & 0.9747 & 0.9971 & 0.9824 & -- & --\\
    & & -- & \qquad$\pm$0.0004 & \qquad$\pm$0.0025 & \qquad$\pm$0.0002 & \qquad$\pm$2.9e-05 & \qquad$\pm$5.0e-06 & \qquad$\pm$1.2e-05 & -- & --\\
    \bottomrule
  \end{tabular}}
\end{table}

\begin{figure}[ht!]
\captionsetup{justification=centering}
\centering
\includegraphics[width=0.3\linewidth]{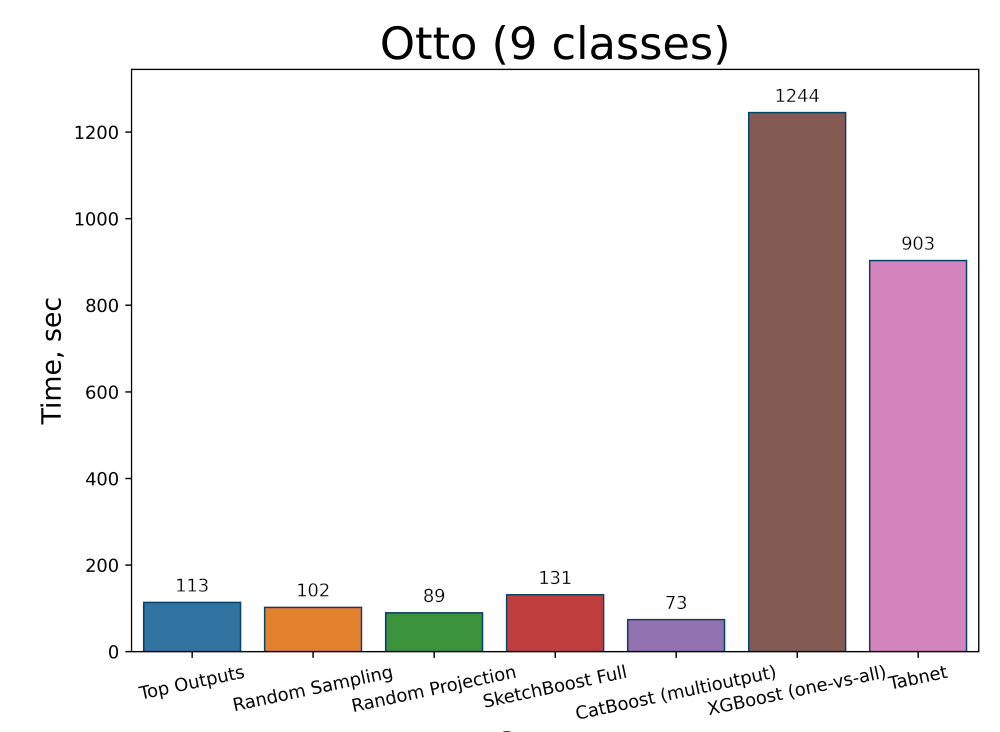}
\includegraphics[width=0.3\linewidth]{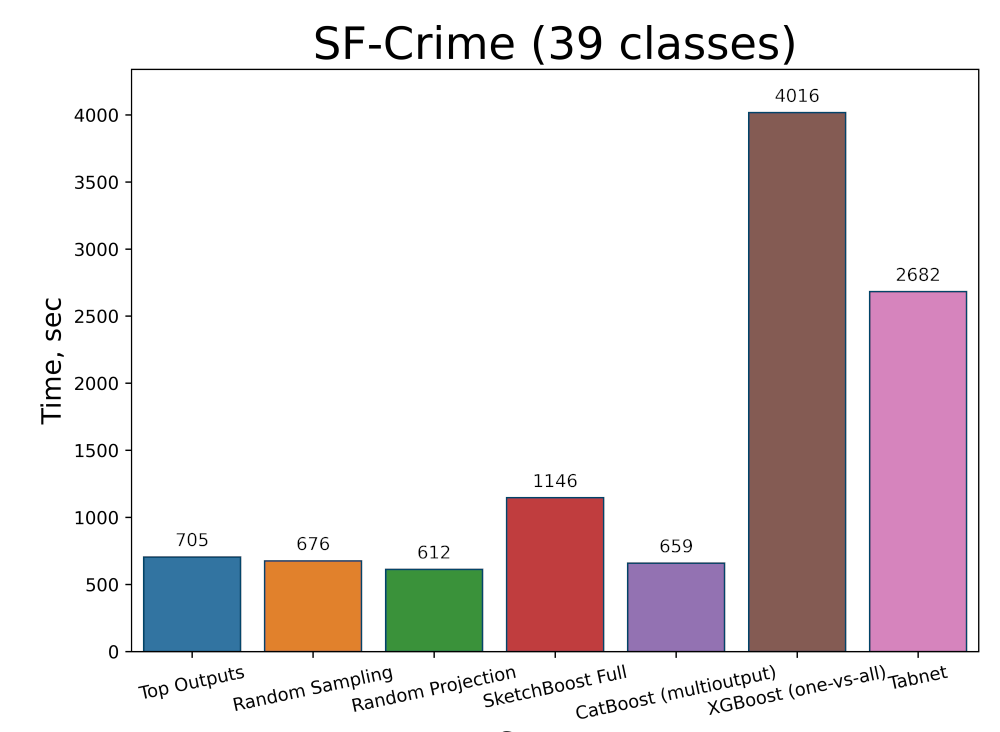}
\includegraphics[width=0.3\linewidth]{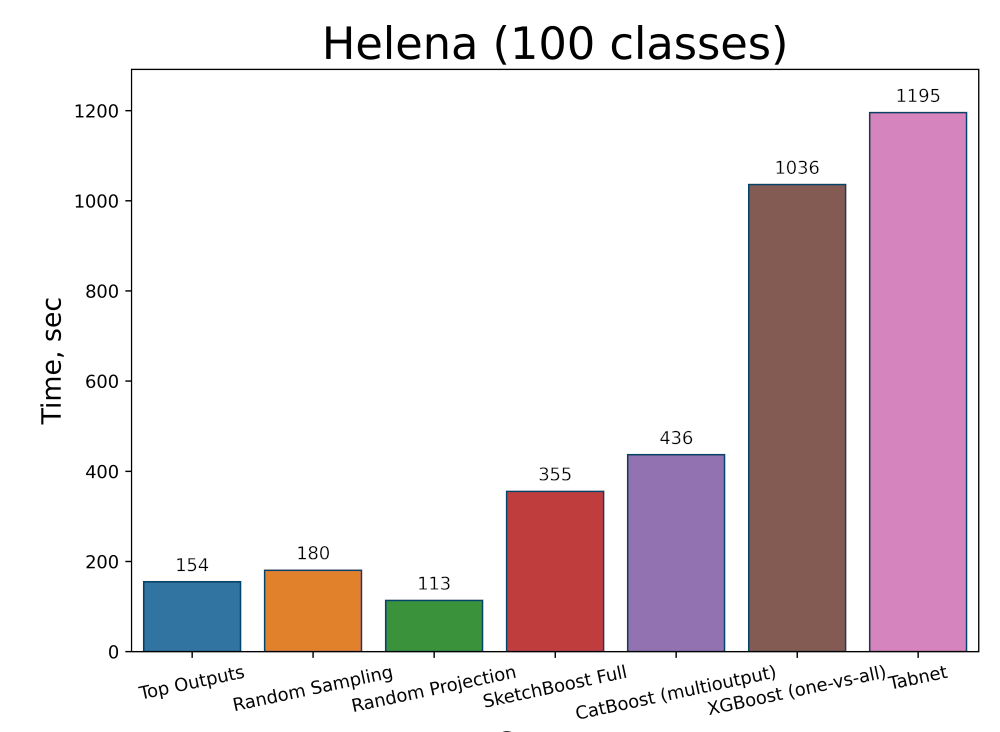}
\\
\includegraphics[width=0.3\linewidth]{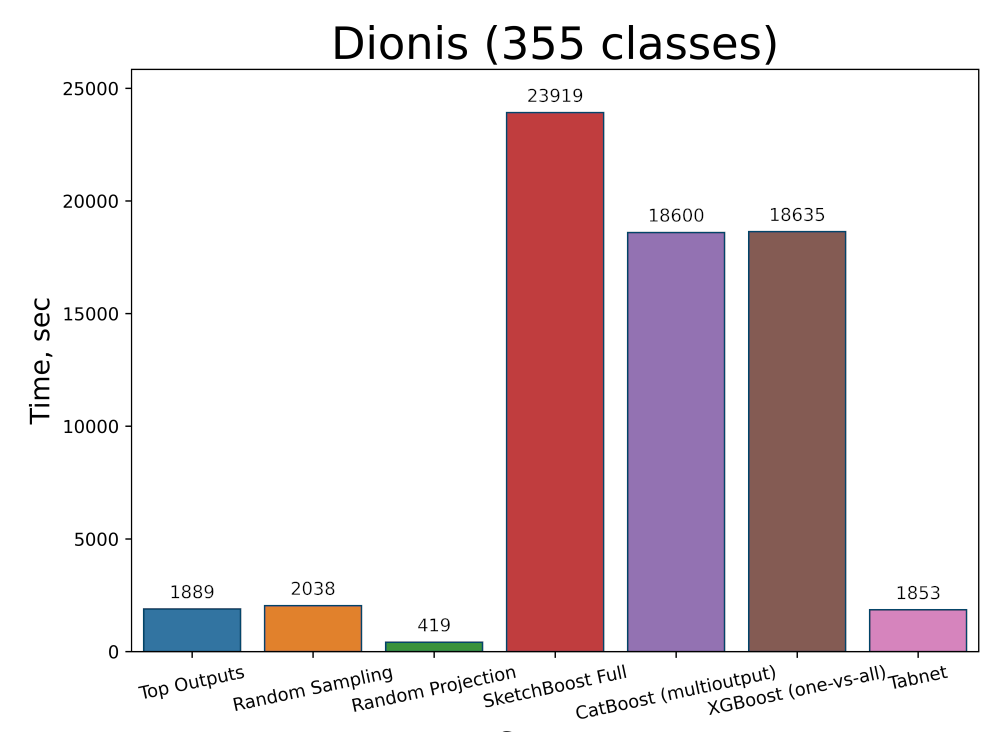}
\includegraphics[width=0.3\linewidth]{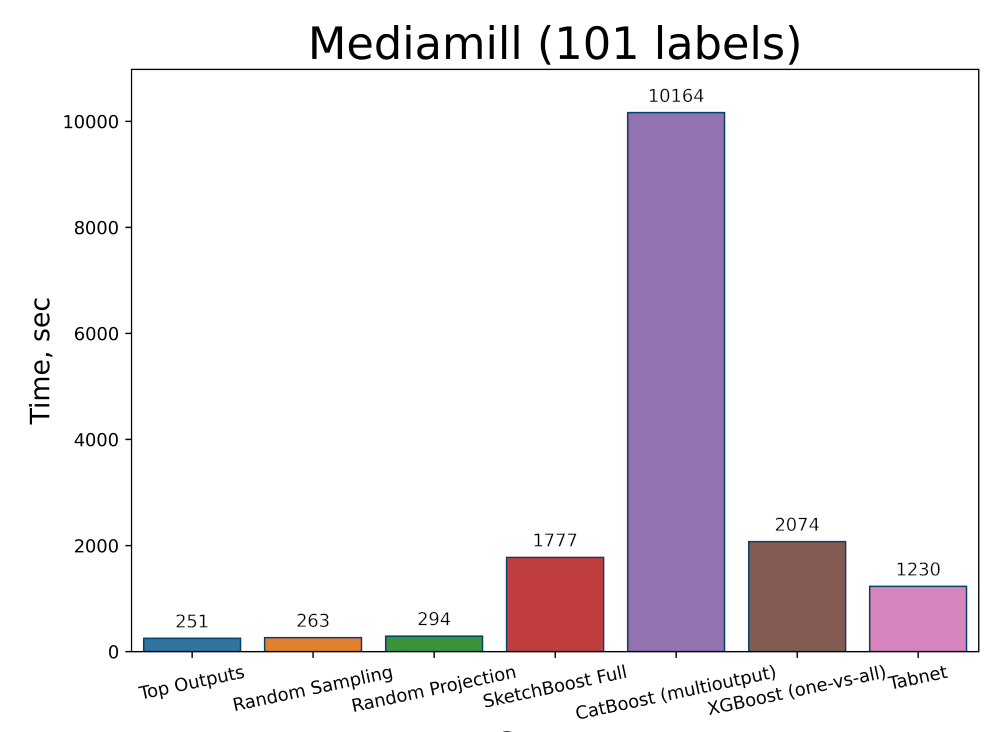}
\includegraphics[width=0.3\linewidth]{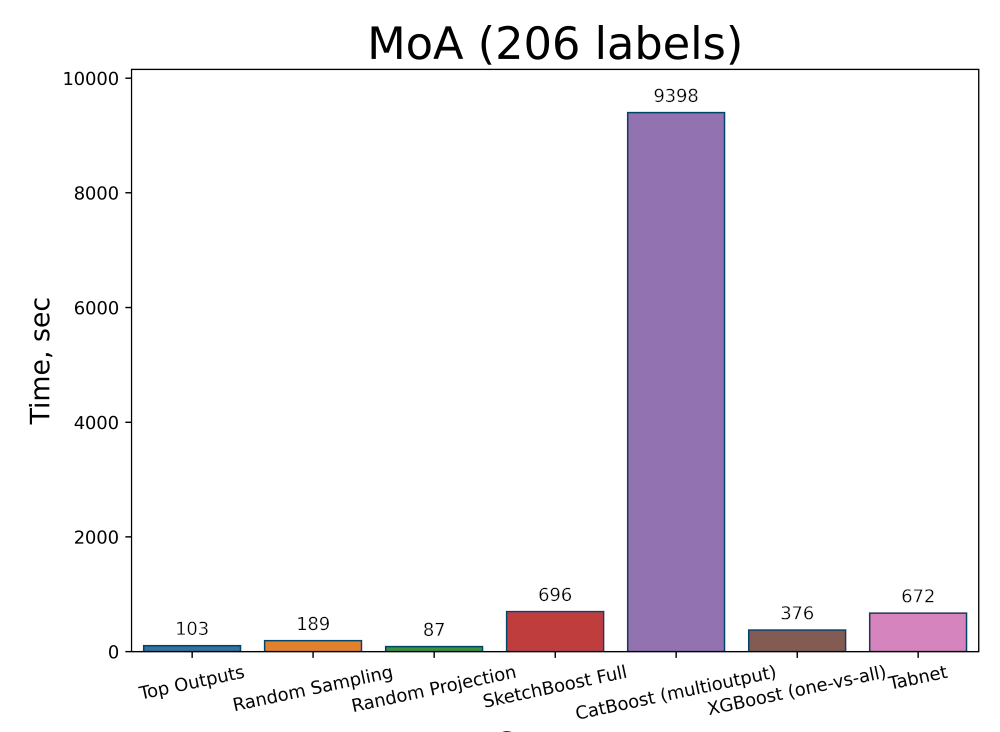}
\\
\includegraphics[width=0.3\linewidth]{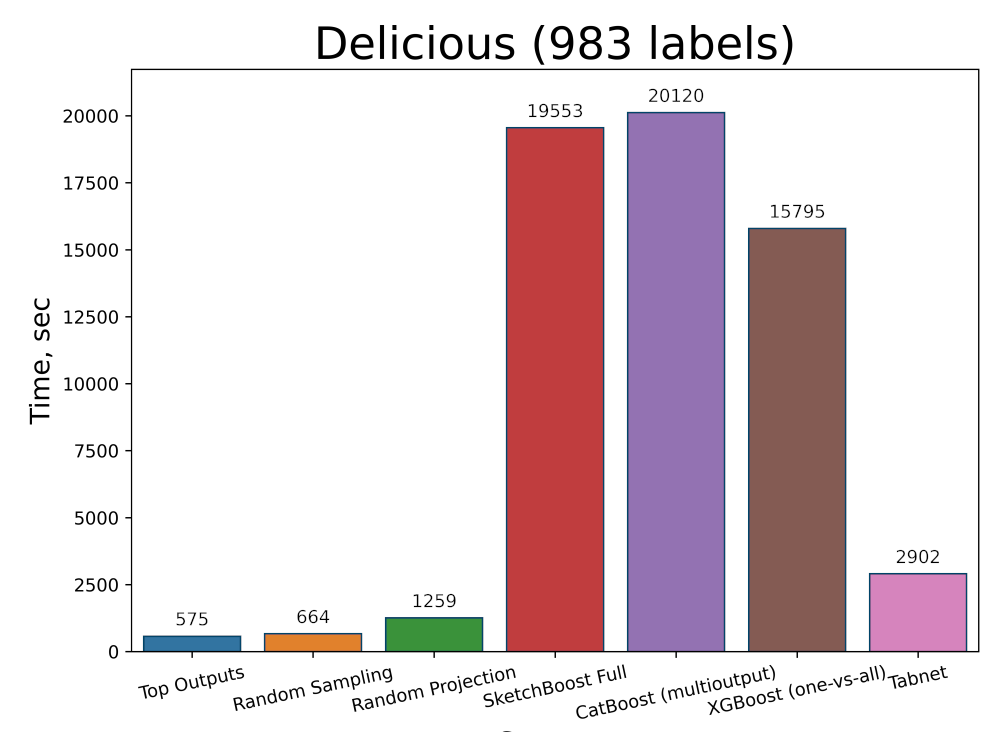}
\includegraphics[width=0.3\linewidth]{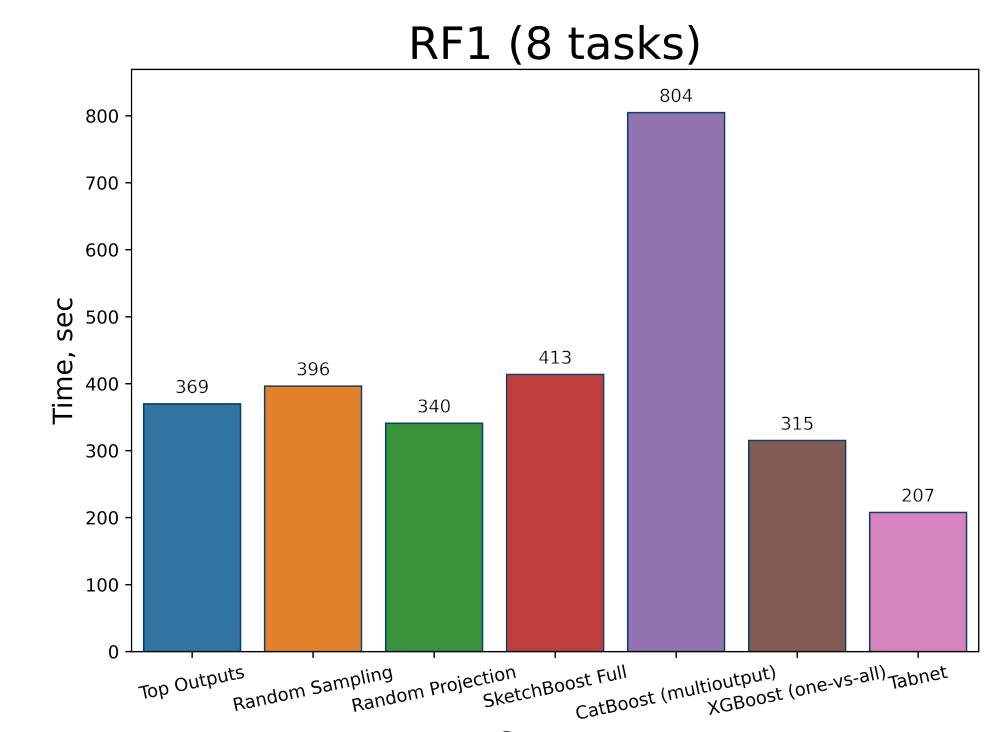}
\includegraphics[width=0.3\linewidth]{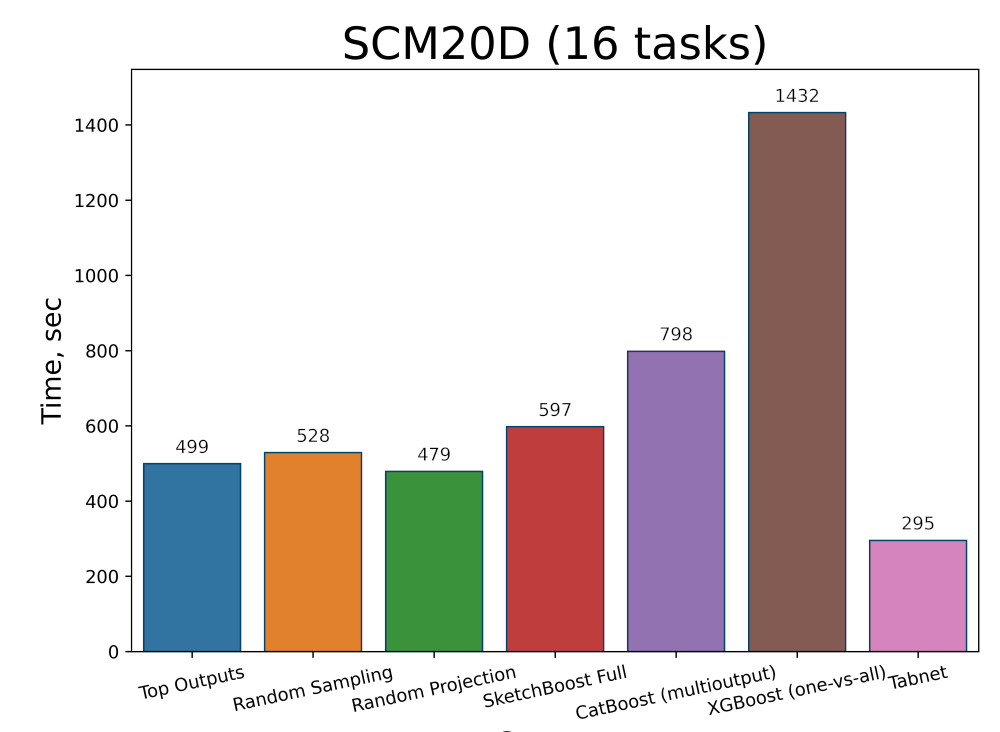}
\\
\caption{ Training time per fold in seconds for the best sketching dimension $\rdim$.\label{pic:train_time}}
\end{figure}

\begin{table}[htbp]
  \setlength\tabcolsep{3pt}
  \setlength\extrarowheight{2pt}
  \captionsetup{justification=centering}
  \centering
  \caption{ 
  Training time per fold in seconds for all sketching dimensions $\rdim$.
  \label{tb:detailed_train_time}}
  \scalebox{0.75}{
  \begin{tabular}{@{\extracolsep{4pt}}llccccccccc@{}}
    \toprule
    & &
    \multicolumn{9}{c}{\textbf{Dataset}} 
    \vspace{0.4em}\\
    \cline{3-11} \vspace{-0.2em}
    \textbf{Algorithm} & \textbf{}  & 
    \textbf{Otto} & \textbf{SF-Crime} & \textbf{Helena} & 
    \textbf{Dionis} & \textbf{Mediamill} & \textbf{MoA} & \textbf{Delicious} & \textbf{RF1} & \textbf{SCM20D}\\
    \midrule
    XGBoost & & 1244 & 4016 & 1036 & 18635 & 2074  & 376 & 15795 & 315 & 1432\\
    CatBoost & & 73 & 659 & 436 & 18600 & 10164 & 9398 & 20120 & 804 & 798 \\
    TabNet & & 903 & 2563 & 1196 & 1853 & 1231 & 672 & 2902 & 207 & 296 \\
    SketchBoost Full  & & 131 & 1146 & 355 & 23919 & 1777 & 696 & 19553 & 413 & 597\\
    \midrule
    Top Outputs & $k=1$ & 129 & 174 & 154 & 783 & 251 & 40 & 213 & 351 & 458 \\
      & $k=2$ & 126 & 207 & 151 & 810 & 276 & 45 & 229 & 364 & 476\\
      & $k=5$ & 113 & 270 & 146 & 1003 & 313 & 59 & 274 & 369 & 499\\
      & $k=10$ & -- & 425 & 138 & 1293 & 386 & 69 & 375 & -- & 551\\
      & $k=20$ & -- & 705 & 156 & 1889 & 529 & 103 & 575 & -- & --\\
    \midrule
    Random Sampling & $k=1$ & 104 & 198 & 180 & 835 & 263 & 61 & 230 & 347 & 485\\
      & $k=2$ & 102 & 219 & 180 & 880 & 273 & 75 & 243 & 354 & 491\\
      & $k=5$ & 116 & 299 & 185 & 1087 & 319 & 104 & 314 & 396 & 528\\
      & $k=10$ & -- & 422 & 198 & 1404 & 399 & 135 & 432 & -- & 590\\
      & $k=20$ & -- & 676 & 213 & 2038 & 559 & 189 & 664 & -- & --\\
    \midrule
    Random Projection & $k=1$ & 89 & 136 & 109 & 419 & 235 & 26 & 212 & 331 & 466\\
      & $k=2$ & 87 & 159 & 113 & 464 & 243 & 29 & 235 & 340 & 479\\
      & $k=5$ & 107 & 233 & 116 & 629 & 294 & 39 & 295 & 393 & 528\\
      & $k=10$ & -- & 365 & 128 & 895 & 369 & 55 & 436 & -- & 594\\
      & $k=20$ & -- & 612 & 149 & 1417 & 527 & 87 & 1259 & -- & --\\
    \bottomrule
  \end{tabular}}
\end{table}

\begin{table}[ht!]
  \setlength\tabcolsep{3pt}
  \setlength\extrarowheight{2pt}
  \captionsetup{justification=centering}
  \centering
  \caption{
  Number of boosting iterations to convergence (for GBDTs).\\{\small (Although the number of iterations for XGBoost is small, it uses the one-vs-all strategy, \\and therefore the actual amount of trees in the ensemble equals this number multiplied by the output size $\odim$.)}\label{tb:best_iter}}
  \scalebox{0.72}{
  \begin{tabular}{@{\extracolsep{4pt}}lcccccc@{}}
    \toprule
    & \multicolumn{4}{c}{\textbf{SketchBoost}} & \multicolumn{2}{c}{\textbf{Baseline}} 
    \vspace{0.4em}\\
    \cline{2-5}  \cline{6-7} \vspace{-0.2em}
    \textbf{Dataset} & \textbf{Top Outputs} & \textbf{Random Sampling} & \textbf{Random Projection} & \textbf{SketchBoost Full} & \textbf{CatBoost} & \textbf{XGBoost} \\
     &\small{(for the best $k$)} & \small{(for the best $k$)} & \small{(for the best $k$)} & \small{(multioutput)} & \small{(multioutput)} & \small{(one-vs-all)}\\
    \midrule
    \textbf{Multiclass classification  } &  & & &  & & \\
    Otto (9 classes)& 4799 & 5424 & 5201 & 4424 & 5534 & 2142\\
    SF-Crime (39 classes) & 3790 & 3726 & 3611 & 3754 & 3993 & 1212\\
    Helena (100 classes) & 15042 & 16975 & 11670 & 13492 & 11238 & 1563\\
    Dionis (355 classes) & 17039 & 17990 & 11509 & 18519 & 19858 & 2681\\
    \midrule
    \textbf{Multilabel classification} &  & & & & & \\
    Mediamill (101 labels) & 18623 & 19961 & 17826 & 17927 & 8983 & 1878\\
    MoA (206 labels) & 2606 & 5542 & 2093 & 2240 & 4239 & 471 \\
    Delicious (983 labels) & 7063 & 8015 & 7541 & 6911 & 3956 & 1611\\
    \midrule
    \textbf{Multitask regression} &  & & & & & \\
    RF1 (8 tasks) & 16102 & 17076 & 16815 & 17001 & 19999 & 19994\\
    SCM20D (16 tasks) & 19992 & 19991 & 19993 & 19992 & 19998 & 19998\\
    \bottomrule
  \end{tabular}}
\end{table}

\clearpage 
\subsection{Comparison with GBDT-MO}
\label{sec:experiment_gbdtmo}

Here we provide details on comparison of 
SketchBoost and CatBoost with 
GBDT-MO and GBDT-MO (sparse) introduced in \citep{gbdt-mo-2021}. The GBDT-MO implementation\footnote{\url{https://github.com/zzd1992/GBDTMO}} 
is evaluated on CPU utilizing 8 threads per run 
(as it was done before).
The experiment design and datasets are taken 
from original paper \citep{gbdt-mo-2021}.
For all the evaluated algorithms,
we use hyperparameters provided in the original paper. 
For GBDT-MO, we use the best sparsity parameter $K$ 
which is also provided in the original paper. 
The only difference in our experiments 
is model training and evaluation 
which is done using 5-fold cross-validation 
(see \Cref{sec:experiment_design}) instead of using the 
test set for both early stopping and performance evaluation
(as was done in the experiments for GBDT-MO\footnote{\url{https://github.com/zzd1992/GBDTMO-EX}}).
We argue the latter method leads to the effect of quality overestimation. We also note that in the original paper
results for GBDT-MO (sparse) are provided
only for 4 datasets out of 6 datasets considered 
(and we also use only these 4 datasets).
As it is done in the original paper, we 
use accuracy as the performance measure.
The experimental results are given below.


\begin{table}[htbp]
  \setlength\tabcolsep{3pt}
  \setlength\extrarowheight{2pt}
  \captionsetup{justification=centering}
  \centering
  \caption{ 
  Comparison with GBDT-MO.
  Test scores (accuracy for classification and \\RMSE for regression) and their standard deviation
  for all sketching dimensions $\rdim$.\label{tb:detailed_test_score_gbdtmo}}
  \scalebox{0.74}{
  \begin{tabular}{@{\extracolsep{4pt}}llcccc@{}}
    \toprule
    & &
    \multicolumn{4}{c}{\textbf{Dataset}} 
    \vspace{0.4em}\\
    \cline{3-6} \vspace{-0.2em}
    \textbf{Algorithm} & \textbf{}  & 
    \textbf{MNIST} & \textbf{Caltech} & \textbf{NUS-WIDE} & 
    \textbf{MNIST-REG} \\
    & & (10 classes) & (101 classes) & (81 labels) & (24 tasks)
    \\
    \midrule
    CatBoost & & 0.9684$\pm$0.004 & 0.5049$\pm$0.0167 & 0.9893$\pm$0.0001 & 0.2708$\pm$0.0023\\
    \midrule
    GBDT-MO Full & & 0.976$\pm$0.004 & 0.4469$\pm$0.059 & 0.9891$\pm$0.0002 & 0.2723$\pm$0.0026\\
    GBDT-MO (sparse) & & 0.9758$\pm$0.0048 & 0.4796$\pm$0.0375 & 0.9892$\pm$0.0006 & 0.2736$\pm$0.0017\\
    \midrule
    SketchBoost Full & & 0.973$\pm$0.0028 & 0.5549$\pm$0.008 & 0.9893$\pm$0.0002 & 0.266$\pm$0.0019\\
    \midrule
    Random Sampling & $k=1$ & 0.973$\pm$0.0045 & 0.5704$\pm$0.0273 & 0.9892$\pm$0.0003 & 0.2671$\pm$0.0011\\
      & $k=2$ & 0.975$\pm$0.0034 & 0.5704$\pm$0.0174 & 0.9891$\pm$0.0003 & 0.2678$\pm$0.0015\\
      & $k=5$ & 0.9755$\pm$0.0042 & 0.5599$\pm$0.0146 & 0.9887$\pm$0.0002 & 0.2671$\pm$0.0012\\
      & $k=10$ & 0.9753$\pm$0.0007 & 0.5623$\pm$0.0165 & 0.989$\pm$0.0002 & 0.2661$\pm$0.0019\\
      & $k=20$ & -- & 0.5691$\pm$0.0127 & 0.9889$\pm$0.0001 & 0.2665$\pm$0.0014\\
    \midrule
    Random Projection & $k=1$ & 0.9737$\pm$0.0023 & 0.5623$\pm$0.0159 & 0.9897$\pm$0.0003 & 0.2657$\pm$0.0018\\
      & $k=2$ & 0.9722$\pm$0.0037 & 0.5537$\pm$0.0064 & 0.9893$\pm$0.0004 & 0.2661$\pm$0.002\\
      & $k=5$ & 0.974$\pm$0.0032 & 0.5605$\pm$0.0137 & 0.9896$\pm$0.0003 & 0.2658$\pm$0.0013\\
      & $k=10$ & 0.9722$\pm$0.0045 & 0.5358$\pm$0.0157 & 0.9893$\pm$0.0004 & 0.2663$\pm$0.0007\\
      & $k=20$ & -- & 0.5488$\pm$0.0332 & 0.9892$\pm$0.0004 & 0.2654$\pm$0.0012\\
    \bottomrule
  \end{tabular}}
\end{table}

\begin{table}[htbp]
  \setlength\tabcolsep{3pt}
  \setlength\extrarowheight{2pt}
  \captionsetup{justification=centering}
  \centering
  \caption{ 
  Comparison with GBDT-MO. Training time per fold in seconds\\
  for all sketching dimensions $\rdim$.
  \label{tb:detailed_train_time_gbdtmo}}
  \scalebox{0.74}{
  \begin{tabular}{@{\extracolsep{4pt}}llcccc@{}}
    \toprule
    & &
    \multicolumn{4}{c}{\textbf{Dataset}} 
    \vspace{0.4em}\\
    \cline{3-6} \vspace{-0.2em}
    \textbf{Algorithm} & \textbf{}  & 
    \textbf{MNIST} & \textbf{Caltech} & \textbf{NUS-WIDE} & 
    \textbf{MNIST-REG} 
    \\
    & & (10 classes) & (101 classes) & (81 labels) & (24 tasks)
    \\
    \midrule
    CatBoost & & 156 & 136 & 13857 & 964\\
    \midrule
    GBDT-MO Full & & 362 & 776 & 2606 & 210\\
    GBDT-MO (sparse) & & 399 & 1312 & 3660 & 163\\
    \midrule
    SketchBoost Full & & 46 & 13 & 87 & 90\\
    \midrule
    Random Sampling & $k=1$ & 66 & 15 & 36 & 110\\
      & $k=2$ & 99 & 42 & 145 & 85\\
      & $k=5$ & 102 & 40 & 148 & 98\\
      & $k=10$ & 88 & 41 & 151 & 120\\
      & $k=20$ & -- & 40 & 158 & 78\\
    \midrule
    Random Projection & $k=1$ & 45 & 16 & 72 & 38\\
      & $k=2$ & 70 & 13 & 71 & 38\\
      & $k=5$ & 66 & 15 & 73 & 51\\
      & $k=10$ & 70 & 14 & 49 & 44\\
      & $k=20$ & -- & 14 & 48 & 45\\
    \bottomrule
  \end{tabular}}
\end{table}

\clearpage

\subsection{Experiment with synthetic dataset}
\label{sec:synthetic_dataset}
The aim of this experiment is to illustrate the dependence of the time cost for training $100$ trees on the number of outputs for popular GBDT frameworks on GPU. To do this, we train each framework twice on each task for $100$ and $200$ iterations and then calculate the difference in time. This allows us to estimate the time costs of $100$ boosting iterations regardless the constant time costs such as features quantization and data transfer. 
\par
In more detail, since our goal is not to measure model quality and since we need alike datasets that vary only in the output dimension, we consider synthetic datasets generated (with the same feature parameters) by the algorithm proposed in \citep{guyon2003design} and implemented in the scikit-learn library (v1.0.2)\footnote{See the  \href{https://scikit-learn.org/stable/modules/generated/sklearn.datasets.make_classification.html}{Scikit-learn documentation}.}. The dataset features are generated with $2000$k rows and $100$ features ($10$ features ar informative, $20$ features are their linear combinations, and others are redundant). At each iteration, the number of classes is changed over the grid 
$\{5,10,25,50,100,250,500\}$.
After the dataset is generated, we compute and report the time difference between $100$ and $200$ iterations for XGBoost, CatBoost, and SketchBoost with Random Projections (sketch dimension $k=5$). The hardware used in this experiment is the same as described in \Cref{sec:experiment_design}. The main hyperparameters are chosen to be similar for all boosting frameworks. Namely, 
(1) trees are grown with depth-wise policy with maximal depth limited to $6$, (2) row and column sampling is disabled,
(3) learning rate is set to $0.01$, and
(4) L2 regularization term is set to $1$, L1 regularization is disabled.

\end{document}